\documentclass[letterpaper]{article} 
\usepackage{aaai2026}  
\usepackage{times}  
\usepackage{helvet}  
\usepackage{courier}  
\usepackage[hyphens]{url}  
\usepackage{graphicx} 
\usepackage{natbib}  
\usepackage{caption} 
\usepackage{amsmath}
\usepackage{amsfonts}
\usepackage{multirow}
\usepackage{makecell}

\usepackage{xcolor}
\UseRawInputEncoding\usepackage{xcolor}
\usepackage{booktabs}
\usepackage{adjustbox}
\usepackage{arydshln}

\newcommand{\answerTODO}[1]{\textcolor{red}{#1}}

\usepackage{algorithm}
\usepackage{algorithmic}
\usepackage{booktabs} 
\usepackage{newfloat}
\usepackage{listings}
\DeclareCaptionStyle{ruled}{labelfont=normalfont,labelsep=colon,strut=off} 
\floatstyle{ruled}
\newfloat{listing}{tb}{lst}{}
\floatname{listing}{Listing}
\title{Beyond a Joke: Multi-Angle Reasoning for Detecting and Explaining Harmful Humor in Memes}
\author{
    Shanhong Liu\textsuperscript{\rm 1},
    Pai Chet Ng\textsuperscript{\rm 2},
    De Wen Soh\textsuperscript{\rm 1},
    Malika Meghjani\textsuperscript{\rm 1},
    Konstantinos N. Plataniotis\textsuperscript{\rm 3}
}

\affiliations{
    \textsuperscript{\rm 1}Singapore University of Technology and Design, Singapore\\
    \textsuperscript{\rm 2}Singapore Institute of Technology, Singapore\\
    \textsuperscript{\rm 3}University of Toronto, Canada\\
    shanhong\_liu@mymail.sutd.edu.sg, paichet.ng@singaporetech.edu.sg,\\
    \{dewen\_soh, malika\_meghjani\}@sutd.edu.sg, 
    kostas@ece.utoronto.ca
}

\usepackage{bibentry}

\begin{document}

\maketitle

\begin{abstract}
Internet memes intertwine visual cues, textual content, and cultural context, making them particularly challenging to interpret in scenarios where humor, sarcasm, and harmful intent coexist. These complexities highlight the need for explainable meme understanding systems that can provide reliable and structured reasoning to support both accurate classification and human interpretability. However, existing multimodal classifiers either overlook such intertwinings or provide only limited interpretability. In this paper, we introduce MAR-12, a novel framework that leverages Vision–Language Models (VLMs) for meme detection and understanding, where both humorous and hateful elements can coexist. The framework first interprets each meme through twelve structured perspectives derived from humor and hate theory. It then applies a role-aware soft-gated attention mechanism to learn how much each perspective should contribute, followed by a prototype-based classifier for final prediction. Finally, explanations were synthesized using both perspective-specific reasoning and learned attention weights, ensuring transparent and context-grounded justifications. We evaluate MAR-12 on the PrideMM and Memotion datasets, where it achieves up to 80.3\% accuracy for humor detection and 75.9\% accuracy for hate detection, outperforming state-of-the-art approaches. Furthermore, both human and GPT-4-based evaluations confirm that MAR-12 produces coherent and persuasive explanations, particularly for memes in which humorous and harmful cues co-occur.

\noindent
\textbf{\emph{Warning: Contains potentially offensive content.}}

\end{abstract}


\noindent
\textbf{CODE} --- \url{https://anonymous.4open.science/r/MAR-12} 

\section{Introduction}

\begin{quote}
\textit{There is a thin line that separates laughter and pain, comedy and tragedy, humor and hurt}.
\begin{flushright}
— Erma Bombeck
\end{flushright}
\end{quote}

Internet memes that blend images with short text have become a pervasive mode of online communication. While many are designed simply to entertain, this same expressive format can also be weaponized: humor often functions as a rhetorical shield for hateful or discriminatory content.
This creates a blurred and often deliberately manipulated boundary between what is perceived as “funny” and what is harmful. During the COVID-19 pandemic, for instance, humorous memes served as coping mechanisms for fear and uncertainty \cite{yus2023meme}, yet similar formats were appropriated to spread racist, conspiratorial, or stigmatizing narratives \citep{steffen2025more}. Likewise, LGBTQ+ memes have long supported community expression and identity \cite{griffin2021living,he2025chinese}, but the same visual-textual patterns have also been used to mock or marginalize queer identities \cite{baker2020putin,crawford2021memes}. These examples underscore the need for automated systems that can differentiate benign humor from harmful messaging, especially when the intended meaning is intentionally ambiguous.

\begin{figure}[t!]
    \centering
    \includegraphics[width=1.05\linewidth, trim={0cm, 1cm, 0cm, 0cm}, clip]{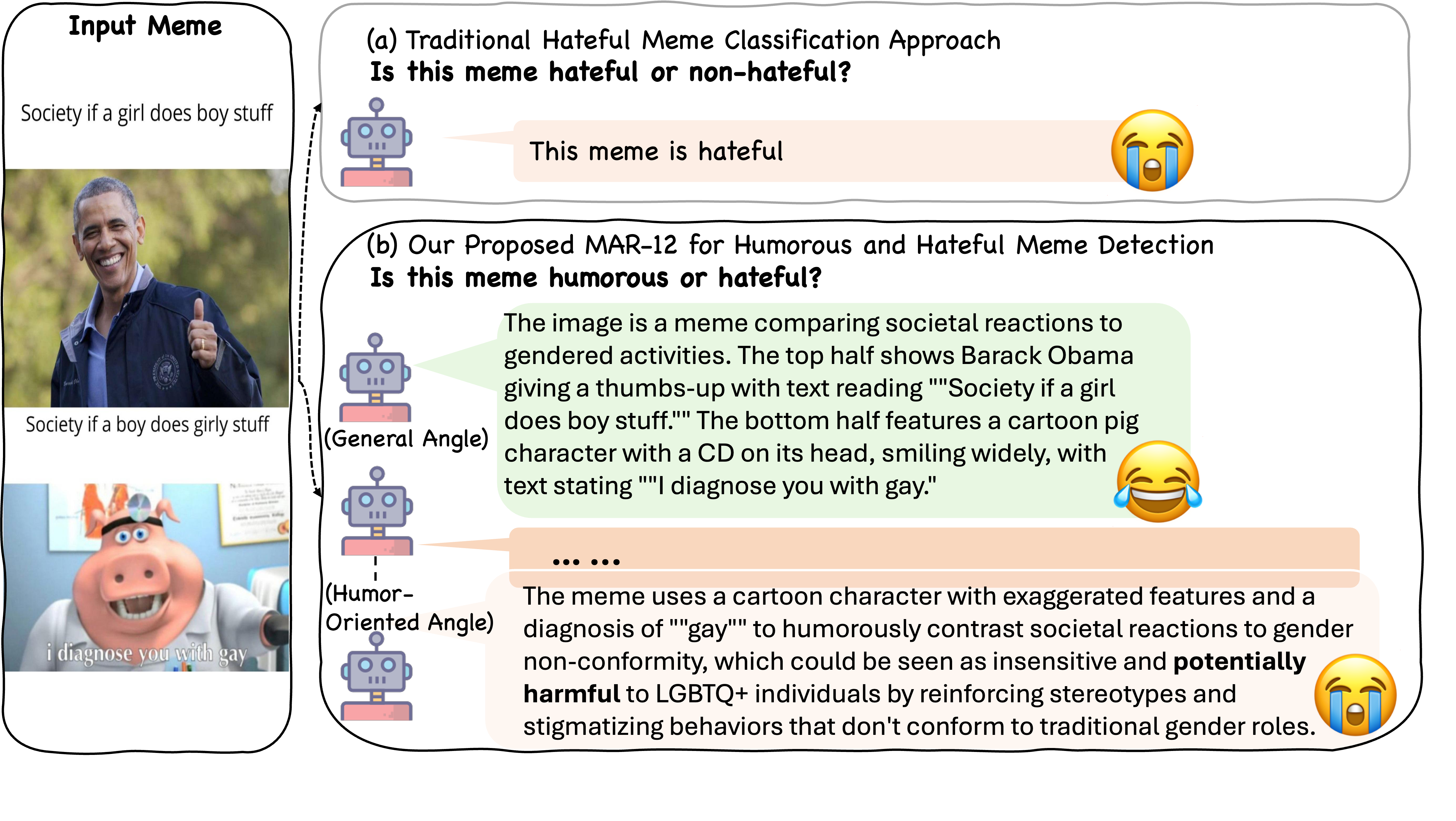}
    \caption{ Comparison of our proposed MAR-12 (bottom) with traditional meme detection (upper).}
    \label{fig:intro}
    \vspace{-0.3cm}
\end{figure}

A central challenge in meme understanding is that humor is inherently subjective and multidimensional: what one viewer finds amusing may appear offensive, confusing, or harmful to another \citep{attardo2024linguistic}. Moreover, humor in memes is not generated by a single cue. It may emerge from visual incongruity, textual irony, emotional mismatch, cultural reference, exaggeration, or linguistic play, often intertwined with undertones of hostility or stereotyping \cite{kalloniatis2024computational}. As illustrated in Figure~\ref{fig:intro}(a), traditional learning frameworks, whether unimodal or multimodal, tend to compress these heterogeneous cues into a single embedding, making them prone to overlooking the subtle interactions that determine whether a meme is humorous, hateful, or both \citep{liu2024sarcasm,shah2024memeclip,pramanick2021momenta}. Although recent work fine-tunes vision-language models (VLMs) to improve alignment between images and text for hateful meme detection \citep{cao2023pro,hee2025contrastive}, these models typically behave as black boxes, providing little insight into why a given meme is categorized as humorous or harmful.

Recent interpretability-oriented approaches generate rationales or stepwise explanations for meme classification \citep{ji2024capalign, rizwan2026see,liu2026yes}, marking an important shift toward transparency. However, these methods generally assume that humor and hate are mutually exclusive or that a single explanatory pathway is sufficient. In reality, many memes deliberately blend comedic devices with harmful insinuations, producing layered meanings that require reasoning across multiple, and often conflicting cues. A meme may be humorous due to incongruity or exaggeration while simultaneously invoking stereotypes or derogatory comparisons. Existing explanation frameworks tend to highlight only the dominant signal and overlook minority cues critical for identifying harmful intent. Moreover, they rarely distinguish foundational content from higher-level interpretive mechanisms, making the resulting explanations difficult to align with established theories of humor and harmful speech. These limitations suggest that understanding humor and hate co-occurrence demands multiple coordinated interpretive lenses rather than one monolithic rationale.

To address this gap, we propose Multiple-Angle Reasoning (MAR), which guides the vision-language model (VLM) to interpret a meme through a coordinated set of twelve complementary perspectives. Instead of collapsing all cues into a single embedding, MAR decomposes interpretation into foundational content (e.g., image description, extracted text), humor mechanisms (e.g., irony, absurdity, affective contrast, linguistic play), and social-meaning cues (e.g., harmfulness, intent), forming a coverage set of reasoning pathways grounded in humor theory and harmful speech analysis. These perspectives are not rigid components; the model is free to emphasize or suppress them depending on the meme. To integrate these outputs, MAR employs a role-aware soft-attention module that learns how each perspective should contribute to the final decision by explicitly encoding its semantic function. Unlike standard attention, which treats inputs as interchangeable, this mechanism preserves functional distinctions and captures how the relevance of different cues shifts across humorous, hateful, or mixed-intent memes. As illustrated in Figure~\ref{fig:intro}(b), this yields an interpretable distribution over reasoning angles and a final prediction that transparently reflects how competing interpretations were weighted.
Our key contributions are:
\begin{itemize}
\item Our proposed MAR is the first framework to jointly examine humor and hate in memes through a diverse set of interpretable angles, covering mechanisms such as visual/textual incongruity, cultural grounding, emotional contrast, absurdity, linguistic play, harmfulness, and inferred intent.
\item We introduce a role-aware soft-attention mechanism that dynamically weights these perspectives based on the input meme, enabling flexible, human-aligned interpretation rather than fixed or hard-coded reasoning.
\item Through extensive experiments across two Meme datasets containing both humorous and hateful content, we demonstrate that MAR consistently outperforms state-of-the-art models on both tasks, particularly in ambiguous cases where humor is intentionally used to mask harmful messaging.
\end{itemize}

\section{Related Works}

\subsection{Meme Detection}
Early work on humorous meme classification relied on multimodal fusion of visual and textual features~\cite{guo2020guoym}. Later models incorporated attention mechanisms to improve cross-modal alignment~\cite{liu2024sarcasm}. However, these approaches still treat humor as a flat label and do not explicitly consider irony, absurdity, affective contrast, exaggeration, linguistic play, that shape humorous intent. They also overlook that humor may coexist with subtle hostility. In parallel, hateful meme detection has advanced through large multimodal benchmark, such as the Hateful Memes Challenge~\cite{kiela2020hateful} and related datasets~\cite{cao2022prompting}, and fine-tuning CLIP  and similar models to improve multimodal semantics~\cite{hee2022explaining,thapa2025multimodal}. These models added object-, face-, or text-level cues to detect explicit harm, yet they still treat harmfulness as a single signal and often fail when hate is masked by humor, satire, or cultural nuance.

On the other hand, we reframe meme understanding as a multiple-angle reasoning problem. Instead of compressing diverse cues into a single embedding, we decompose interpretation into twelve complementary perspectives, enabling the model to differentiate benign humor from ambiguous or veiled hostility, something prior methods struggle with. Through a role-aware soft-attention mechanism, MAR-12 learns which perspectives are most informative for humorous, hateful, or mixed-intent memes, yielding flexible, transparent, and human-aligned predictions. This holistic treatment directly mitigates the shortcomings of unimodal, multimodal, and black-box VLM-based systems by capturing the nuanced interplay between humor and harm.

\subsection{Reasoning-Based Meme Understanding}
Recent work increasingly uses LLM- and VLM-based approaches to enhance interpretability in harmful or ambiguous memes\cite{hee2025demystifying, rizwan2025exploring}. Few-shot and zero-shot systems such as \cite{rizwan2025exploring} and MiND \cite{liu2025mind}, LoReHM \cite{huang2024towards}, and related reasoning-oriented methods \cite{lin2025ask} employ visual–textual entailment or chain-of-thought prompting to explain why a meme may be harmful. Debate-inspired frameworks like ExplainHM \cite{lin2024towards} simulate opposing viewpoints and use a judge model to consolidate arguments. Although generative reasoning approaches show promise, they remain constrained by structural assumptions: debate-style systems reduce interpretation to a binary contest, while single- or dual-chain explanations cannot capture the multidimensional space needed to model humor and hate co-occurrence, affective contrast, cultural grounding, or linguistic play.

Rather than relying on one or two reasoning trajectories, our MAR-12 encourage the VLM to reflect on a meme through twelve complementary perspectives spanning foundational content, humor mechanisms, and social-meaning cues. This structured decomposition avoids the binary framing of debate-style methods and better captures the diverse ways humor and harm interact. A role-aware soft-attention mechanism further learns which perspectives matter most for each input, enabling flexible, context-sensitive explanations that highlight ambiguity, competing cues, and mixed intent. By modeling humorous and harmful signals in a unified and interpretable manner, our approach produces explanations that are more faithful, nuanced, and aligned with human interpretive processes than prior generative reasoning systems.

 \begin{figure*}[t!]
    \centering
    \includegraphics[width=.85\linewidth, trim={0cm, 1.2cm, 0.3cm, 0cm}, clip]{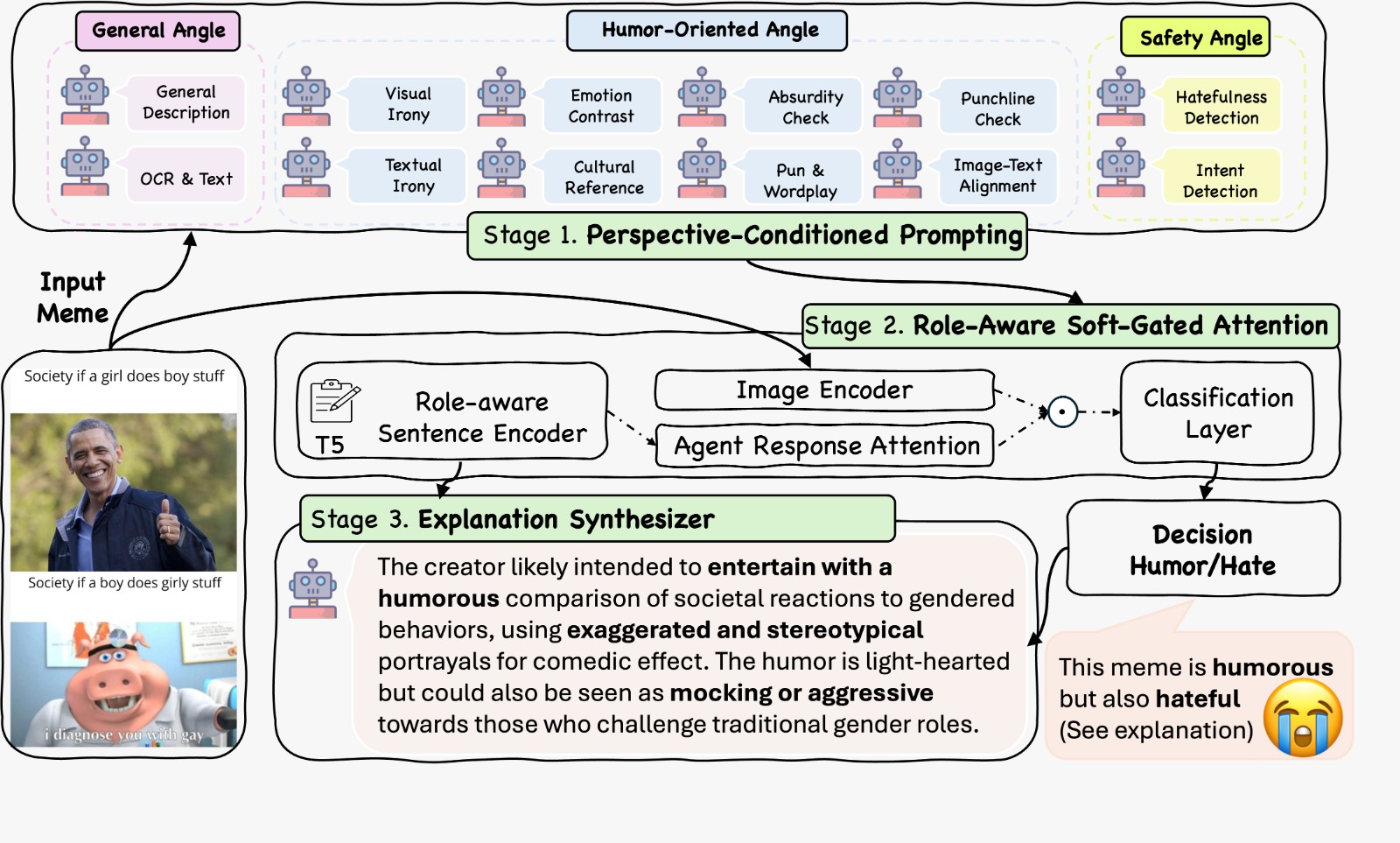}  
    \caption{Overview of the proposed Multiple-Angle Reasoning (MAR) framework. Stage 1 Perspective-Conditioned Prompting. Stage 2 Role-Aware Soft-Gated Attention. Stage 3 Explanation via LLM-based Judemeny.}
    \label{fig:mar12}
\end{figure*}

\section{Methodology}
\label{sec:method}
To model the complex and often conflicting interpretive space involved in meme understanding, we introduce MAR-12, a framework that encourages a VLM to reflect on a meme through a structured set of complementary perspectives.
Rather than forcing the model into fixed behaviors, MAR-12 acts as a form of structured cognitive scaffolding, analogous to how a human might be prompted to “look at it from another angle” or “put yourself in someone else’s shoes.” These perspectives provide breadth, while a role-aware soft-gated attention mechanism determines which angles matter most for each meme, ensuring freedom in reasoning without hard-coding outcomes.
As illustrated in Figure~\ref{fig:mar12}, MAR-12 consists of three stages:
(1) \emph{Perspective-conditioned prompting},
(2) \emph{Role-aware soft-gated attention}, and
(3) \emph{Explanation via LLM-Based Judgment}.

\subsection{Perspective-Conditioned Prompting}

Given a meme \( m = (I, T) \) consisting of an image \( I \) and associated text \( T \), MAR-12 interprets it through a structured set of twelve perspectives \( \mathcal{P} = \{p_1, \dots, p_{12}\} \). Each perspective corresponds to a distinct interpretive lens grounded in linguistic humor theory, affective contrast modeling, and harmful-speech analysis. 


\paragraph{Role-Conditioned Prompting.}
To elicit each reasoning angle, we prompt the VLM by assigning it a perspective-specific role, analogous to asking a human annotator to temporarily adopt a certain interpretive mindset (“consider the cultural context”, “analyze irony”, “evaluate harmful intent”).
Formally, the response for perspective \( p_i \) generates a natural-language reasoning sequence, i.e., 
\begin{equation}
    r_i = \mathcal{Q}_\theta(I, T, p_i),
\end{equation}
where $r_i$ is a natural-language reasoning sequence specific to viewpoint \( p_i \).
Collecting all twelve responses yields:
\begin{equation}
    \mathcal{R}_m = \{ r_1, r_2, \dots, r_{12} \}.
\end{equation}
Each \( r_i \) provides a viewpoint-specific interpretation, and together they form a diverse basis for downstream aggregation. This stage does not impose constraints on which perspectives the model should rely on; instead, it encourages intellectual breadth while allowing Stage 2 to determine relevance.
Due to space constraints, the full prompts for all twelve perspectives are provided in Appendix~\ref{app:prompting} (see Table~\ref{tab:agent-prompts-appendix}). The appendix describes the rationale, structure, and exact wording of each prompt, allowing readers to fully reproduce our perspective-conditioned prompting setup.

\subsection{Role-Aware Multi-Angle Fusion}
To integrate the twelve reasoning traces, we employ a role-aware soft-gated attention mechanism that differentiates between the functional roles of each perspective (e.g., humor vs. safety). Each reasoning sequence $r_i$ is encoded into a fixed-dimensional embedding $\mathbf{h}_i \in \mathbb{R}^d$. 
Unlike standard attention which treats inputs as interchangeable, we introduce a role embedding $\mathbf{e}_i \in \mathbb{R}^d$ that captures the specific conceptual function of perspective $p_i$.
The relevance score $u_i$ is computed by integrating the reasoning content with its functional role:
\begin{equation}
    u_i = \mathbf{w}^\top \tanh(\mathbf{W}_h \mathbf{h}_i + \mathbf{W}_e \mathbf{e}_i + \mathbf{b}),
\end{equation}
where \( \mathbf{W}_h \), \( \mathbf{W}_e \), \( \mathbf{w} \), and \( \mathbf{b} \) are learnable parameters. 
The normalized weights $\alpha_i$ ensure the contribution of each perspective is context-dependent, yielding an aggregated representation 
\begin{equation}
    \mathbf{h}_{\text{agg}} = \sum_{i=1}^{12} \alpha_i \mathbf{h}_i.
\end{equation}
This representation is then concatenated with the visual embedding $\mathbf{v}$ to form the final multimodal vector $\mathbf{z} = [\mathbf{h}_{\text{agg}} \parallel \mathbf{v}']$, which is passed to a prototype-based classifier. 

\paragraph{Prototype-Based Classification. }
To promote geometric interpretability in decision-making, We employ a cosine prototype-based decision function, which is computed as
\begin{equation}
    \hat{y} = \arg\max_k \ \cos(\mathbf{z}, \mathbf{c}_k).
\end{equation}
where each class \( k \in \{1,2\} \) (humor vs.\ non-humor or hate vs.\ non-hate) is represented by a learnable prototype \( \mathbf{c}_k \).

\subsection{Explanation Synthesizer}

Beyond producing a binary prediction, MAR-12 generates a final explanatory rationale that articulates \emph{why} the model arrived at its decision. This step is essential for transparency, especially in cases where humorous cues and harmful intent intersect in subtle or conflicting ways. To construct this explanation, we repurpose the VLM as an explanation synthesizer. The model is prompted with three key pieces of information: the predicted label \( \hat{y} \), the set of multi-angle reasoning outputs \( \mathcal{R}_m = \{r_1, \dots, r_{12}\} \), and the learned attention weights \( \{\alpha_i\}_{i=1}^{12} \) that indicate the relevance of each perspective.

To ensure that the explanation is grounded in the same reasoning process used during classification, MAR-12 provides VLM with a structured instruction prompt that explicitly references both the content of each reasoning angle and its learned importance. 
The model is guided using a template of the following form:
\begin{quote} \small
\textit{
``You are given a meme classification result. The final prediction is: [LABEL].  
Below are the interpretations of the meme from multiple reasoning angles, each accompanied by its relevance weight:  
[REASONING\_TEXT + WEIGHTS].  
Based on these weighted perspectives, summarize the key factors that led to the decision. Explain which perspectives were most influential, which were less relevant, and how they collectively justify the final classification.''}
\end{quote}

Formally, the explanation is produced as
\begin{equation}
    \text{Explanation} = \mathcal{Q}^{\text{exp}}_\theta(\hat{y}, \mathcal{R}_m, \{\alpha_i\}),
\end{equation}
where the concatenated, weight-conditioned reasoning signals serve as inputs to the explanation module. This design ensures that the explanation meaningfully reflects the same multi-angle reasoning used by the classifier, rather than being a generic or post-hoc justification.
The resulting explanation highlights the dominant interpretive angles driving the prediction, acknowledges conflicting or low-weight cues when necessary, and provides a human-readable account of how humor or harmful meaning was inferred. This final stage ensures that MAR-12 functions not merely as a classifier but as a transparent reasoning system whose decisions can be examined and understood. By conditioning the synthesis on the same role-aware attention used by the classifier, MAR-12 achieves process-level faithfulness, where the explanation remains a transparent account of the model's internal evidence flow even in cases of misclassification. 


\begin{table}[h!]
\centering
\caption{Statistical distributions of datasets where ``Hu" and ``Ha" represent humor and hateful, ``Non-Hu" and ``Non-Ha" represent non-humor and non-hateful}
\small
\resizebox{\linewidth}{!}{
\begin{tabular}{l|l|lcc}
\toprule
{Task} & {Split} & {Label} & {PrideMM} & {Memotion} \\
\midrule
\multirow{4}{*}{Humor}
& \multirow{2}{*}{Train} &  Hu         & 2944 & 4272 \\
&                        &  Non-Hu     & 1385 & 1321 \\
& \multirow{2}{*}{Test}  &  Hu        & 318  & 1069 \\
&                        &  Non-Hu      & 190  & 330 \\
\midrule
\multirow{4}{*}{Hate}
& \multirow{2}{*}{Train} &  Ha           & 2121 & 3423 \\
&                        &  Non-Ha       & 2208 & 2170 \\
& \multirow{2}{*}{Test}  &  Ha           & 248  & 856  \\
&                        &  Non-Ha        & 260  & 543  \\
\bottomrule
\end{tabular}}
\label{tab:dataset-stats}
\end{table}

\begin{table}[h!]
\caption{Comparison of LLaVA  and Qwen-VL on humor classification with and without few-shot reasoning.}
\begin{tabular}{lccc}
\toprule
\text{Model} & \text{ACC (\%)} & \text{AUROC (\%)} & \text{\(\Delta\) ACC (\%)} \\
\midrule
LLaVA                       & 67.46 & 65.75 & -- \\
L-Reasoning   & 63.12 & 56.69 & $-$4.34 \\
\midrule
Qwen-VL                     & 48.72 & 57.41 & -- \\
Q-Reasoning & 58.97 & 60.44 & +10.25 \\
\bottomrule
\end{tabular} 
\label{tab:llm-comparison}
\end{table}


\section{Experiments Settings}

\begin{table*}[t!]
\centering
\caption{Performance comparison across humor and hate meme detection tasks for PrideMM and Memotion datasets.}
\begin{tabular}{l|l|ccc|ccc}
\toprule
\multirow{2}{*}{\text{Task}} & \multirow{2}{*}{\text{Model}} & \multicolumn{3}{c|}{\text{PrideMM}} & \multicolumn{3}{c}{\text{Memotion}} \\
\cmidrule(lr){3-5} \cmidrule(lr){6-8}
 &  & \text{ACC (\%)} & \text{AUC (\%)} & \text{F1 (\%)} & \text{ACC (\%)} & \text{AUC (\%)} & \text{F1 (\%)} \\
\midrule
\multirow{7}{*}{\text{Humor}} 
& Visual Only (Resnet50 + MLP)       & 66.08    & 58.01    & 61.67     & 76.48    & 50.57    & 66.88\\
& Text Only  (T5 + MLP)      & 67.85    & 62.38    & 66.10     &  76.27  &  50.64    & 67.02   \\
& MemeCLIP \citep{shah2024memeclip} & 78.30  & 73.27  & 76.99   & 76.34  & 51.31  & 67.77\\
& MOMENTA \citep{pramanick2021momenta}& 73.57  & 65.79  & 69.92   & 76.48  & 50.00$^{\dagger}$
& 66.50\\
& PromptHate \citep{cao2022prompting}& 73.77  & 71.01  & 73.46   & 76.48  & 50.00$^{\dagger}$  & 66.19 \\
& LoReHM (LLaVA-34B) \citep{huang2024towards} & 70.09  & 56.71  & 64.07   & 76.18  & 50.91  & 67.32\\
& MiND (Qwen2.5-VL-32B ) \citep{liu2025mind}  & 54.45 & 51.00 & 50.43   & 60.72  & 52.39  & 53.27\\
\cdashline{2-8}
& \text MAR-12 (T5 Embedding)      & 79.68    & 75.11    & 78.63     & 79.10    & 74.51    & 77.85 \\
& \text MAR-12 (Clip Embedding)      & 80.08    & \textbf{77.64}    & \textbf{79.82}   & \textbf{79.85}    & \textbf{77.04}    & \textbf{78.88} \\
& \text MAR-12 (T5xClip Embedding)      & \textbf{80.28}    & 76.00    & 79.37     & 79.15    & 75.84    & 78.38 \\
\midrule
\multirow{7}{*}{\text{Hate}} 
& Visual Only  (Resnet50 + MLP)      & 62.72   & 62.89   & 62.57     & 61.47     & 50.00$^{\dagger}$
& 46.80 \\
& Text Only   (T5 + MLP)        & 72.78   & 72.77   & 72.78     &59.61      & 51.95   &  54.46   \\ 
& MemeCLIP \citep{shah2024memeclip}          & \textbf{75.35} & 75.09 & \textbf{75.35}   & 61.76  & 50.99  & 49.58\\
& MOMENTA \citep{pramanick2021momenta}            & 75.15 & 75.22 & 75.14   & 61.47  & 50.00$^{\dagger}$
& 46.80\\
& PromptHate \citep{cao2022prompting}        & 71.80 & 71.66 & 71.70   & 61.47  & 50.00$^{\dagger}$ 
& 46.81\\
& LoReHM (LLaVA-34B) \citep{huang2024towards}  & 65.05 & 65.16 & 64.95   & 53.72  & 52.43  & 54.13\\
& MiND (Qwen2.5-VL-32B ) \citep{liu2025mind}  &61.73  & 60.87 & 61.65   & 73.05  & 68.46  & 70.59 \\
\cdashline{2-8}
& \text MAR-12 (T5 Embedding)       & \textbf{75.35}   & \textbf{75.36}   & \textbf{75.35}     & \textbf{76.15}     & 70.20   & 71.12\\
& \text MAR-12 (Clip Embedding)      & 73.37    & 73.81    & 72.66     & 75.53    & 73.42    & 71.05 \\
& \text MAR-12 (T5xClip Embedding)      & 73.18    & 73.47    & 72.86     & 75.89    & \textbf{74.04}    & \textbf{72.00} \\  
\bottomrule
\end{tabular}
\small{$^{\dagger}$\textit{An AUC of 50.00 alongside accuracy levels matching the majority class (e.g., 76.48\% for humor or 61.47\% for hate) indicates that the baseline collapsed to predicting the majority label for all samples, thus failing to learn discriminative rank-ordering.}}
\label{tab:unified_results}
\end{table*}

\subsection{Evaluation Datasets}
To evaluate MAR-12 in settings where humor and harmful meaning frequently interact, we require datasets that contain both humor and hate annotations for the same meme. However such datasets are extremely scarce: among existing meme benchmarks, only two publicly available resources, i.e., PrideMM \cite{shah2024memeclip} and Memotion \cite{sharma2020semeval}, provide dual labels that allow a model to study humor and hate co-occurrence on the same image–text pair. This makes them uniquely suited for evaluating multi-angle reasoning approaches such as MAR-12, where the goal is to disentangle humorous mechanisms from potentially harmful or offensive undertones.

\emph{PrideMM} focuses on LGBTQ related memes, many of which combine satire, wordplay, and cultural references with socially or politically charged commentary. This mixture often produces content where humor can either soften or conceal hostile messaging, making it a strong testbed for evaluating MAR-12's ability to analyze competing interpretive cues. 

\emph{Memotion} provides meme annotations across humor, sarcasm, and offensiveness. Its broader thematic coverage introduces a diverse range of humorous styles, absurd, sarcastic, situational, and text-dominant, while also capturing offensiveness that may emerge independently or in parallel with humor. For experimentation, we binarize humor by grouping \emph{funny}, \emph{very funny}, and \emph{hilarious} as humorous and \emph{not funny} as non-humorous. Offensive labels are treated as hateful.

\begin{table}[t!]
\centering
\caption{Summary of baseline models and our proposed MAR-12 method, with total and trainable parameter counts.}
\small
\resizebox{\linewidth}{!}{
\begin{tabular}{lll}
\toprule
\textbf{Model} & \textbf{Total Parameters} & \textbf{Trainable Parameters} \\
\midrule
ResNet-50 + MLP (Visual Baseline)       & 25.48M    & 1.97M    \\
T5 + MLP (Text Baseline)                & 110.02M   & 110.02M  \\
PromptHate                              & 355.41M   & 355.41M  \\
MemeCLIP                                & 431.29M   & 3.68M    \\
MOMENTA                                 & 431.29M   & 3.68M    \\
LoReHM                                  & 34.00B    & 34.00B        \\
\textbf{MAR-12 (Ours)}                  & 1.31M     & 1.31M    \\
\bottomrule
\end{tabular}}
\label{tab:model_summ}
\end{table}

\subsection{Models}
We compare MAR-12 against three families of strong baselines to demonstrate its advantage over unimodal, multimodal, and reasoning-based approaches.  
\textbf{(i) Unimodal baselines} evaluate the contribution of isolated modalities: a Visual-Only model (ResNet50 + MLP) and a Text-Only model (T5 + MLP).  
\textbf{(ii) Multimodal CLIP-style baselines} include \textbf{MOMENTA} \cite{pramanick2021momenta}, which enriches CLIP features with object- and face-level cues from VGG-19 and textual features from DistilBERT, ~\textbf{MemeCLIP} \cite{shah2024memeclip},  which incorporates trainable adapters into CLIP for multi-aspect meme classification while preserving its generalization capabilities, and  ~\textbf{PromptHate}\cite{cao2022prompting}, which reformulates meme classification as masked language modeling with prompt templates using RoBERTa.
\textbf{(iii) Reasoning-based VLM baselines} include \textbf{LoReHM} \cite{huang2024towards}, which uses a single LLM agent to generate reasoning chains through visual-textual entailment and commonsense inference, and \textbf{MiND} \cite{liu2025mind}, which incorporate prompting or chain-of-thought reasoning but do not structure reasoning across diverse interpretive angles as MAR-12 does.  
This grouping highlights that MAR-12 improves over systems designed around single cues, multimodal matching, and generic reasoning alike.

As summarized in Table~\ref{tab:model_summ}, MAR-12 is designed to be an extremely lightweight framework, containing only 1.31 million trainable parameters. This is orders of magnitude smaller than competitive reasoning-based baselines like LoReHM, which relies on a 34.00 billion parameter Large Multimodal Model. 
This efficiency is achieved by keeping the heavy VLM and feature encoders frozen, ocusing the learning process solely on the role-aware attention mechanism and the prototype-based classification head. Consequently, MAR-12 provides a high-performance alternative to end-to-end fine-tuning, offering significant depth in reasoning without the prohibitive computational costs and memory requirements associated with large-scale model optimization.


\subsection{Implementation Details}
\paragraph{Vision Language Models.} 
Before evaluating MAR-12 directly, we first assess which VLM is most responsive to multi-angle reasoning prompts. We compare LLaVA \cite{liu2023visual} and Qwen-VL~\cite{bai2023qwen} under two settings: (1) direct classification with a single prompt and (2) classification augmented with few-shot reasoning exemplars. In the few-shot setting, each exemplar includes a meme image, its label, and a GPT-4\cite{achiam2023gpt} generated explanation, allowing us to examine whether the model can integrate external reasoning cues. As shown in Table~\ref{tab:llm-comparison}, LLaVA performs best in the direct setting (67.46\% ACC) but degrades when reasoning exemplars are introduced, suggesting difficulty in leveraging structured interpretive signals. Qwen-VL, however, improves from 48.72\% to 58.97\% accuracy and from 57.41\% to 60.44\% AUC, demonstrating a better performance in internalizing structured reasoning. 
These results support our choice of Qwen-VL as the backbone for MAR-12, which relies heavily on high-quality reasoning across multiple interpretive perspectives.

\paragraph{MAR-12 Encoders.} We use CLIP ViT‑B/32 as the vision encoder, with image inputs resized to 224×224 pixels. The agent text embeddings were obtained using T5 (or CLIP text encoder in some variants) and projected into a 1024‑dimensional mapped space from the original 768‑dimensional embeddings. The fusion module included 1 mapping layer and 1 pre‑output layer, with dropout probabilities of 0.1, 0.4, and 0.2 applied at different stages. The classification head employed a cosine‑similarity‑based prototype classifier with a scale of 30 and margin ratio of 0.2.

\paragraph{MAR-12 Training and Efficiency.} 
We trained MAR-12 on a high-performance computing server equipped with NVIDIA L40S GPUs (46,068 MiB VRAM) and dual AMD EPYC 7543 32-Core Processors. To ensure computational efficiency and scalability for large-scale meme moderation, we utilized mixed precision training (torch.float16), which significantly optimized memory utilization and reduced the overall training footprint.
We used 4 data‑loading workers and fixed the random seed to 42 for reproducibility. 
We used a learning rate of $1 \times 10^{-4}$, weight decay of $1 \times 10^{-4}$, and a batch size of 32 to train MAR-12 over 20 epochs.  
Training the model for 20 epochs on the Memotion dataset took less than 30 minutes, with an average inference time of 3.5 seconds per meme. 
This inference latency accounts for the parallelized processing of the twelve VLM reasoning traces and the final synthesis by the explainer.
A comprehensive breakdown of the computing resources, software environment (CUDA 12.5), and specific training configurations is available in the Appendix.
As for the evaluation of the models' performance, we base our choices on the average  of  their Accuracy  (Acc.), Area  Under  the  Receiver  Operating  Characteristics  curve(AUROC), and F1-score.  We  optimize these models using Adam optimizer \cite{kingma2014adam} and  are  implemented  in  PyTorch  using  the  Huggingface's \texttt{Transformers} ~\footnote {https://huggingface.co/docs/transformers} library.

\section{Experiments}
\subsection{Performance Comparison with Baselines}

After selecting Qwen-VL as the backbone, we evaluate MAR-12 against unimodal, multimodal, and reasoning-based baselines. As shown in Table~\ref{tab:unified_results}, MAR-12 achieves state-of-the-art results across both datasets and tasks. For humor detection on PrideMM, MAR-12 (CLIP) attains 80.08\% ACC, 77.64\% AUC, and 79.82\% F1, with the T5+CLIP fusion variant slightly improving accuracy further. On Memotion, MAR-12 again outperforms all baselines, reaching 79.85\% ACC and 78.88\% F1. Similar improvements appear in the hate detection task: MAR-12 (T5) matches or exceeds the strongest baselines on PrideMM with the highest AUC (75.36\%), and MAR-12 (T5xCLIP) achieves the best overall performance on Memotion (74.04\% AUC and 72.00\% F1). These results demonstrate that MAR-12 surpasses unimodal systems, multimodal CLIP-style models, and reasoning-based VLMs alike. 
Notably, several baselines such as MOMENTA and PromptHate exhibit a performance collapse on the Memotion dataset, achieving high accuracy (~76\%) but random-level AUC (50.00). Our audit confirmed this was due to the models defaulting to majority-class prediction as a result of the dataset's class imbalance (see Table 1). MAR-12 successfully overcomes this limitation by leveraging multi-angle reasoning to maintain high discriminative power (77.04\% AUC on humor)


\subsection{Ablation Studies}
To better understand the contribution of multi-angle reasoning and the role-aware attention mechanism within MAR-12, we conduct two ablation studies. The first evaluates the importance of different perspective groups, while the second examines how various attention mechanisms influence the aggregation of perspective-specific reasoning. 

\paragraph{Effect of Removing Perspective Groups.}  
MAR-12 organizes its twelve reasoning angles into three functional categories: a \emph{humor perspective group} (capturing irony, absurdity, wordplay, and comedic framing), a \emph{safety perspective group} (capturing harmfulness, intent, stereotyping, and derogatory cues), and a \emph{foundational perspective group} (capturing OCR text, literal description, and basic semantic grounding). Table~\ref{tab:ablation1} shows how removing the humor or safety groups affects performance. For humor classification, MAR-12 with all perspectives achieves 79.85\% accuracy and 77.64\% AUC. However, removing either the humor or safety perspectives causes AUC to collapse to 50.00\%, equivalent to random guessing. A similar pattern occurs in the hate task: full MAR-12 obtains 75.53\% accuracy and 73.42\% AUC, but removing humor or safety perspectives again reduces AUC to approximately 50\%.

To ensure a rigorous evaluation of each functional category, we completely retrained the Stage 2 role-aware attention module and prototype-based classifier from scratch for each ablation setting. We did not merely mask the inputs at inference time; instead, the model was forced to attempt to learn a discriminative decision boundary using only the remaining perspective groups. The collapse of the AUROC to 50.00 (random guessing) when either the humor or safety groups are removed demonstrates that these specialized reasoning signals are indispensable for the learning process. This confirms that foundational perspectives alone are insufficient for capturing the nuanced interplay of humor and harm in memes.
These findings indicate that when key interpretive groups are removed, the model is forced to rely solely on the foundational perspectives, which provide surface-level cues but lack the task-specific discriminative information needed for complex reasoning about humor or harmful intent. Without the humor or safety groups, the role-aware attention cannot meaningfully differentiate between perspectives and thus defaults to majority-class predictions. The near-identical drops observed when removing either group also highlight that humor and safety cues function complementarily in detecting humor and hate co-occurrence, reinforcing the need for a multi-angle interpretive structure.

\paragraph{Effect of Attention Mechanisms.}  
We next evaluate whether MAR-12's improvements stem from the use of role-aware attention or whether simpler attention variants suffice. Table~\ref{tab:ablation2} compares three alternatives, standard soft attention, gated attention, and multi-head attention, using both MLP and ResNet visual backbones. Soft attention achieves moderate performance but shows instability across backbones, as it treats all reasoning embeddings as interchangeable and cannot account for the conceptual distinctions between humor, safety, and foundational perspectives. Gated attention introduces additional nonlinearity but still struggles to encode differences tied to perspective roles, yielding only marginal improvements. Multi-head attention performs worst, particularly with the MLP backbone, as its distributed heads tend to diffuse focus across many perspectives, preventing stable alignment with role semantics.

\begin{table}[t!]
\centering
\caption{Ablation study evaluating MAR-12 when individual perspective groups are removed.}
\begin{tabular}{l|l|cc}
\toprule
\text{Task} & \text{Setting} & \text{ACC (\%)} & \text{AUC (\%)} \\
\midrule
\multirow{3}{*}{\text{Hu}}   
    & MAR-12        & 79.85 & 77.64 \\
    & w/o Humor Perspectives           & 76.41 & 50.00 \\
    & w/o Safety Perspectives          & 76.41 & 50.00 \\
\midrule
\multirow{3}{*}{\text{Ha}}   
    & MAR-12        & 75.53 & 73.42 \\
    & w/o Humor Perspectives           & 61.47 & 50.03 \\
    & w/o Safety Perspectives          & 61.47 & 50.03 \\
\bottomrule
\end{tabular}
\label{tab:ablation1}
\vspace{-0.3cm}
\end{table}

\begin{table}[t!]
\centering
\caption{Ablation study on the role of attention and backbone architecture in MAR-12.}
\resizebox{\linewidth}{!}{
\begin{tabular}{l|cc}
\toprule
\multirow{2}{*}{\text{Method}}   &Humor  & Hate\\
&ACC(\%)  & ACC(\%) \\
\midrule
MLP+Soft Attention      & 75.74 & 70.79 \\
ResNet+Soft Attention   & 74.16 & 67.11 \\
MLP+Gated Attention      & 76.73 & 71.9 \\
ResNet+Gated Attention      & 76.92 & 72.58 \\
MLP+Multi-head Attention        & 72.98 & 66.90 \\
ResNet+Multi-head Attention     & 74.95 & 70.27 \\
MLP+Role-aware Attention (ours)       & \text{79.36} & \text{75.11} \\
ResNet+Role-aware Attention (ours)    & \text{78.78} & \text{75.27} \\
\bottomrule
\end{tabular} }
\label{tab:ablation2}
\vspace{-0.3cm}
\end{table}

In contrast, MAR-12's role-aware attention achieves the highest and most consistent performance across both tasks and backbones. By explicitly incorporating learned role embeddings for each perspective group, the mechanism preserves functional distinctions between different types of reasoning and enables the model to prioritize complementary cues. This prevents the diffusion of attention seen in multi-head variants and avoids the homogenization of embeddings seen in soft attention. With these role distinctions encoded, MAR-12 effectively leverages the diverse reasoning signals produced by the perspective-conditioned prompts, leading to superior generalization and more robust interpretive alignment across datasets.

\begin{table}[t!]
\centering
\caption{Automatic GPT-4 evaluation of the explanation quality of humor and hate tasks on Memotion test sets.}
\label{tab:evaluation}
\begin{tabular}{l|cc|cc}
\toprule
\multirow{2}{*}{\text{Task}} & \multicolumn{2}{c|}{\text{Ha}} & \multicolumn{2}{c}{\text{Non-Ha}} \\
\cmidrule(lr){2-3} \cmidrule(lr){4-5}
 & \text{Hu} & \text{Non-Hu} & \text{Hu} & \text{Non-Hu} \\
\midrule
Informativeness     & 3.38 & 3.46 & 3.81 & 3.79 \\
Readability         & 3.96 & 3.97 & 4.11 & 4.05 \\
Soundness           & 3.64 & 3.79 & 3.76 & 3.69 \\
Conciseness         & 3.90 & 3.92 & 3.92 & 3.94 \\
Persuasiveness      & 3.79 & 3.82 & 3.86 & 3.84 \\
\bottomrule
\end{tabular}
\label{tab:auto}
\vspace{-0.3cm}
\end{table}

\begin{table}[h!]
\centering
\caption{Human evaluation of the explanation quality of humor and hate tasks on Memotion test sets.}
\label{tab:evaluation2}
\begin{tabular}{l|cc|cc}
\toprule
\multirow{2}{*}{\text{Task}} & \multicolumn{2}{c|}{\text {Ha}} & \multicolumn{2}{c}{\text{Non-Ha}} \\
\cmidrule(lr){2-3} \cmidrule(lr){4-5}
 & \text{Hu} & \text{Non-Hu} & \text{Hu} & \text{Non-Hu} \\
\midrule
Informativeness     & 4.03 & 4.02 & 3.88 & 3.75 \\
Readability         & 3.78 & 3.73 & 3.65 & 3.47 \\
Soundness           & 3.73 & 3.67 & 3.63 & 3.48 \\
Conciseness         & 3.88 & 4.12 & 4.00 & 3.85 \\
Persuasiveness      & 3.80 & 3.73 & 3.68 & 3.77 \\
\bottomrule
\end{tabular}
\label{tab:human}
\end{table}
\section{Empirical Analysis}
Since MAR-12's core contribution is interpretable multi-angle reasoning, we evaluate the quality of its explanations using both automatic and human judgments. Explanations in this setting must integrate diverse reasoning perspectives, reflect the learned importance weights, and articulate why humor or hate was inferred from a multimodal meme. We therefore assess not only linguistic fluency but also whether the explanations meaningfully reflect MAR1-2's structured interpretive process.

\begin{figure*}[t!]
    \centering
    \includegraphics[width=1\linewidth]{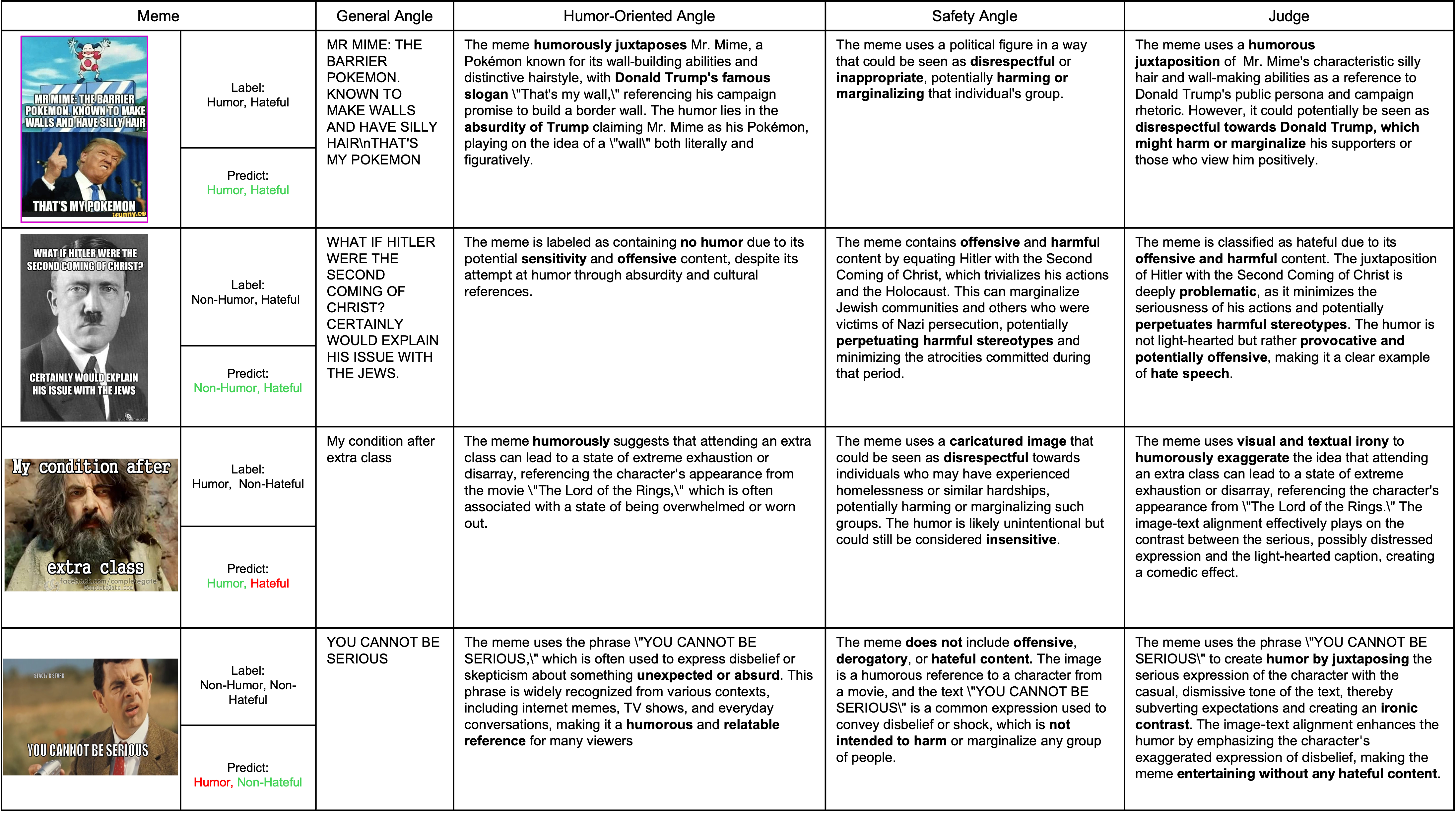}
    \caption{Example memes from four scenarios: (a) hateful and humorous, (b) hateful but non-humorous, (c) non-hateful but humorous, and (d) non-hateful and non-humorous. Correct (in \textcolor{green}{Green}) and incorrect  (in \textcolor{red}{Red}) predictions of our method. Important information is in \textbf{Bold}.}
    \label{fig:error_analysis}
    \vspace{-0.3cm}
\end{figure*}

\subsection{Evaluation of Explainability}
\paragraph{Automatic Evaluation.}  
Following recent practice \cite{lin2024towards}, we use GPT-4 as an automatic evaluator and score each explanation along five criteria that relate directly to MAR-12's design principles. \emph{Informativeness} captures whether the explanation introduces additional contextual or cultural insight beyond surface-level observations, reflecting the breadth of MAR-12's multi-angle reasoning. \emph{Readability} assesses fluency and structural coherence. \emph{Soundness} evaluates whether the explanation presents a logically grounded interpretation consistent with the role-aware attention mechanism. \emph{Conciseness} measures the model’s ability to summarize relevant reasoning without redundancy. Finally, \emph{Persuasiveness} reflects whether the explanation forms a compelling justification aligned with the aggregated reasoning signals. Each criterion is rated on a 5-point Likert scale.

To examine how MAR-12 handles different humor–hate interactions, we evaluate explanations across four interpretive regimes: hateful and humorous, hateful but non-humorous, non-hateful but humorous, and neither hateful nor humorous. Table~\ref{tab:auto} reports averaged GPT-4 scores for each category. We observe that explanations remain consistently strong across most criteria, particularly in Readability (3.96–4.11) and Conciseness (3.90–3.94), indicating that MAR-12 produces structurally coherent and succinct justifications. Soundness is similarly stable (3.64–3.79), suggesting that role-aware attention effectively integrates multiple reasoning angles into a coherent interpretive thread. 
This stability is further supported by our quantitative alignment analysis, which proves the synthesizer's narrative is causally grounded in the model's internal attention weights rather than being a post-hoc justification of the final label.
The most challenging regime is hateful-and-humorous memes, which score lowest on Informativeness (3.38). These cases often require additional cultural or contextual grounding that may not be explicit in the meme, reflecting the inherent difficulty of reconciling competing humorous and harmful cues.

\paragraph{Human Evaluation.}  
Because automatic evaluation cannot fully capture the subjectivity and cultural nuance involved in meme interpretation \cite{lin2024towards}, we complement it with a human evaluation conducted by five English-proficient  university  students using the same five criteria. Table~\ref{tab:human} summarizes the results. Inter-annotator agreement is moderate (ICC = 0.618; Spearman’s $\rho$ = 0.643), consistent with the subjective nature of humor and harm perception. Human judgments reveal patterns similar to the automatic evaluation: explanations for hateful-but-non-humorous memes receive the highest Soundness and Conciseness scores, likely because harmful cues tend to be explicit. Explanations for humorous-but-non-hateful memes achieve the best Readability and Persuasiveness scores, suggesting that MAR-12's humor-focused reasoning is perceived as fluent and coherent. Across all criteria, memes containing both humor and hate exhibit slightly lower scores, further confirming the interpretive difficulty of cases involving ambiguous or dual communicative intent.

Overall, both automatic and human evaluations indicate that MAR-12 produces explanations that are coherent, grounded, and well-aligned with its multi-angle reasoning framework. The model reliably synthesizes diverse interpretive cues into concise and persuasive justifications. The most challenging cases involve humor and hate co-occurrence, where cultural or contextual subtleties may exceed what is explicitly represented in the meme. These findings highlight the importance of multi-angle reasoning: even when faced with conflicting signals, MAR-12 maintains stable explanatory quality while making explicit the interpretive tensions that shape the final prediction.

\begin{table}[t!]
\centering
\caption{Pearson correlations. Alignment captures faithfulness to agent reasoning/attention rather than correctness, whereas margin correlates with correctness as expected.}
\label{tab:faithfulness}
\begin{tabular}{l r}
\toprule
\textbf{Pair} & \textbf{Pearson $r$} \\
\midrule
$\mathrm{align_{humour}}$ vs.\ $\mathrm{hum_{correct}}$ & $0.030$ \\
$\mathrm{align_{hate}}$ vs.\ $\mathrm{hate_{correct}}$ & $-0.026$ \\
$\mathrm{margin_{humour}}$ vs.\ $\mathrm{hum_{correct}}$ & $0.241$ \\
$\mathrm{margin_{hate}}$ vs.\ $\mathrm{hate_{correct}}$ & $0.267$ \\
\midrule
$\mathrm{align_{humour}}$ vs.\ $\mathrm{align_{hate}}$ & $0.425$ \\
$\mathrm{margin_{humour}}$ vs.\ $\mathrm{margin_{hate}}$ & $-0.053$ \\
$\mathrm{align_{humour}}$ vs.\ $\mathrm{margin_{humour}}$ & $-0.045$ \\
$\mathrm{align_{hate}}$ vs.\ $\mathrm{margin_{hate}}$ & $-0.033$ \\
\bottomrule
\end{tabular}
\end{table}
\subsection{Quantitative Faithfulness}
To move beyond perceived quality, we conduct a quantitative analysis of explanation faithfulness
with an Alignment metric that measures the textual grounding of the Explanation Synthesizer in the twelve weighted reasoning traces, weighted by the model's learned attention.
As shown in Table~\ref{tab:faithfulness}, our analysis reveals that faithfulness is uncorrelated with prediction accuracy ($r=0.030$ for humor; $r = - 0.026$ for hate).
The results in Table~\ref{tab:faithfulness} verifies that the synthesizer faithfully narrates the internal evidential path even when the classifier is incorrect. 

Furthermore, the consistency of alignment values across tasks ($r=0.425$) indicates that MAR-12 induces a stable, role-aware explanatory pattern that tracks internal attention rather than inventing its own saliency to match a label. Extended analysis is included in Appendix~\ref{app:judge-faithfulness}.

\subsection{Case Study Analysis}
To illustrate how MAR-12 navigates competing humorous and harmful cues, we analyze four representative memes spanning the full humor and hate interpretive space, as shown in Figure~\ref{fig:error_analysis}. These examples highlight how multi-angle reasoning and perspective-specific attention weights \( \{\alpha_i\} \) shape the final predictions, particularly in cases involving ambiguous communicative intent.

From the natural-language explanations produced by MAR-12, we observe two recurring patterns.  
\text{Firstly}, the coexistence of humor and hate substantially increases classification difficulty. MAR-12 must determine whether humor is benign, whether it masks harmful messaging, or whether it amplifies derogatory undertones. This tension is evident in Figure~\ref{fig:error_analysis}(a), where humor-focused perspectives emphasize the absurd juxtaposition of Mr. Mime, a Pokémon known for creating invisible walls, with the slogan “That’s my wall,” referencing Donald Trump’s political rhetoric. In contrast, safety-focused perspectives attend to the derogatory implications embedded in the comparison. The learned weights \( \alpha_i \) must reconcile these conflicting cues; when they lean toward humorous framing, the model favors a benign interpretation, but when safety cues dominate, it signals harmful intent. A similar conflict occurs in Figure~\ref{fig:error_analysis}(c), where the humorous exaggeration of exhaustion is correctly identified by humor perspectives, yet safety perspectives overweight hardship-related visual elements, leading the model to infer harmful undertones despite the meme being non-hateful.

\text{Secondly}, misclassifications often arise from the inherently subjective and culturally dependent nature of humor and hate annotations. Memes involving sensitive themes, gender, race, religion, or political ideology, can elicit divergent interpretations across annotators. This is reflected in Figure~\ref{fig:error_analysis}(b) and \ref{fig:error_analysis}(d). Although annotators labeled both memes as humorous but non-hateful, MAR-12 predicted them as humorous but hateful. In Figure~\ref{fig:error_analysis}(b), safety-focused perspectives correctly detect the alignment between a historical dictator and a sensitive textual reference, capturing a legitimate basis for concern. In Figure~\ref{fig:error_analysis}(d), the model recognizes the situational humor conveyed by facial expression and caption, yet elevated attention to socio-political cues leads to a cautious interpretation that diverges from annotator consensus. These cases demonstrate how small shifts in \( \alpha_i \) can alter interpretive emphasis, especially when memes rely on subtle social nuance or culturally specific humor.

Taken together, these examples reveal the real-world interpretive complexity of memes where humor and harmful meaning intersect. They also demonstrate a multi-angle, role-aware reasoning framework like MAR-12 can expose the sources of ambiguity, articulate competing interpretations, and justify its final decision in a transparent manner.

\section{Conclusion}
We present MAR-12, a multi-angle reasoning framework for the joint detection and explanation of humorous and hateful content in multimodal memes. By decomposing meme understanding into twelve theory-driven analytical perspectives, our approach addresses the intertwined challenges of performance and interpretability in meme moderation. Through extensive experiments on the PrideMM and Memotion datasets, we demonstrate that MAR-12 consistently outperforms existing multimodal and LLM-based baselines in both detection accuracy and explanation quality. In addition, our human and automatic evaluations show that MAR-12 produces coherent, context-grounded justifications, particularly for challenging cases where humor, sarcasm, and harmful intent co-occur.

\section{Limitations and Future Work} \label{sec:limitations}
There are several directions in which this work can be further improved: 1) MAR-12 relies on predefined analytical perspectives derived from humor and hate theory. While these perspectives capture many common cues, they may not fully represent the diversity of cultural references, regional slang, or evolving meme conventions found online. Expanding the perspective set or incorporating culturally adaptive reasoning modules could improve coverage across different social and linguistic contexts. 2) The current design relies on single-turn explanation synthesis. In some challenging cases, particularly those involving subtle cultural references or sensitive attributes such as race or identity, the generated explanations may miss important contextual details. Future research could explore multi-turn reasoning strategies for explanation generation and incorporating iterative refinement to improve the completeness and reliability of explanations. 3) Although we employ both human evaluation and GPT-4-based automatic assessment to measure explanation quality, discrepancies remain between machine-based and human judgments. For instance, large language models may exhibit systematic biases when evaluating explanations generated by similar models. More robust automatic evaluation protocols, along with larger-scale and more diverse human studies, are needed to provide more reliable and unbiased assessment of explanation quality. 

\section{Ethics and Broader Impact} \label{sec:ethics}
By providing interpretable justifications, MAR-12 aims to empower moderators and researchers to better identify harmful intent, particularly when humor is used to mask hateful or harmful content. Nevertheless, we acknowledge the risk that generated explanations could be misused to craft more persuasive harmful memes or to evade automated moderation systems. We strongly discourage and condemn such actions. We sincerely appreciate all the volunteer annotators and remain mindful of the risks associated with annotation tasks. In response, we (1) informed annotators in advance about the potentially harmful nature of the content, (2) ensured their explicit acknowledgment before participation, and (3) advised them to stop if they felt overwhelmed.

\bibliography{aaai2026}

@incollection{yus2023meme,
  title={Meme-Mediated Humorous Communication},
  author={Yus, Francisco},
  booktitle={Pragmatics of Internet Humour},
  pages={245--307},
  year={2023},
  publisher={Springer}}

@book{attardo2024linguistic,
  title={Linguistic theories of humor},
  author={Attardo, Salvatore},
  volume={1},
  year={2024},
  publisher={Walter de Gruyter GmbH \& Co KG}
}

@article{griffin2021living,
  title={Living through it: Anger, laughter, and internet memes in dark times},
  author={Griffin, Hollis},
  journal={International Journal of Cultural Studies},
  volume={24},
  number={3},
  pages={381--397},
  year={2021},
  publisher={SAGE Publications Sage UK: London, England}
}

@article{he2025chinese,
  title={Chinese medicine as a cure for gayness: Satire as countercultural resistance against heteronormative symbolic violence in digital public sphere},
  author={He, Renyi and Chang, Jiang},
  journal={International Journal of Cultural Studies},
  pages={13678779241308289},
  year={2025},
  publisher={SAGE Publications Sage UK: London, England}
}

@article{baker2020putin,
  title={Putin as gay icon? Memes as a tactic in Russian LGBT+ activism},
  author={Baker, James E and Clancy, Kelly A and Clancy, Benjamin},
  journal={LGBTQ+ activism in Central and Eastern Europe: Resistance, representation and identity},
  pages={209--233},
  year={2020},
  publisher={Springer}
}

@inproceedings{hee2025demystifying,
  title={Demystifying hateful content: Leveraging large multimodal models for hateful meme detection with explainable decisions},
  author={Hee, Ming Shan and Lee, Roy Ka-Wei},
  booktitle={Proceedings of the International AAAI Conference on Web and Social Media},
  volume={19},
  pages={774--785},
  year={2025}
}

@article{kalloniatis2024computational,
  title={Computational humor recognition: a systematic literature review},
  author={Kalloniatis, Antonios and Adamidis, Panagiotis},
  journal={Artificial Intelligence Review},
  volume={58},
  number={2},
  pages={43},
  year={2024},
  publisher={Springer}
}

@article{kiela2020hateful,
  title={The hateful memes challenge: Detecting hate speech in multimodal memes},
  author={Kiela, Douwe and Firooz, Hamed and Mohan, Aravind and Goswami, Vedanuj and Singh, Amanpreet and Ringshia, Pratik and Testuggine, Davide},
  journal={Advances in neural information processing systems},
  volume={33},
  pages={2611--2624},
  year={2020}
}

@inproceedings{cao2023pro,
  title={Pro-cap: Leveraging a frozen vision-language model for hateful meme detection},
  author={Cao, Rui and Hee, Ming Shan and Kuek, Adriel and Chong, Wen-Haw and Lee, Roy Ka-Wei and Jiang, Jing},
  booktitle={Proceedings of the 31st ACM International Conference on Multimedia},
  pages={5244--5252},
  year={2023}
}

@inproceedings{huang2024towards,
  title={Towards Low-Resource Harmful Meme Detection with LMM Agents},
  author={Huang, Jianzhao and Lin, Hongzhan and Ziyan, Liu and Luo, Ziyang and Chen, Guang and Ma, Jing},
  booktitle={Proceedings of the 2024 Conference on Empirical Methods in Natural Language Processing},
  pages={2269--2293},
  year={2024}
}

@article{sharma2020semeval,
  title={SemEval-2020 Task 8: Memotion Analysis--The Visuo-Lingual Metaphor!},
  author={Sharma, Chhavi and Bhageria, Deepesh and Scott, William and Pykl, Srinivas and Das, Amitava and Chakraborty, Tanmoy and Pulabaigari, Viswanath and Gamback, Bjorn},
  journal={arXiv preprint arXiv:2008.03781},
  year={2020}
}

@inproceedings{cao2022prompting,
  title={Prompting for Multimodal Hateful Meme Classification},
  author={Cao, Rui and Lee, Roy Ka-Wei and Chong, Wen-Haw and Jiang, Jing},
  booktitle={Proceedings of the 2022 Conference on Empirical Methods in Natural Language Processing},
  pages={321--332},
  year={2022}
}

@inproceedings{lin2024towards,
  title={Towards explainable harmful meme detection through multimodal debate between large language models},
  author={Lin, Hongzhan and Luo, Ziyang and Gao, Wei and Ma, Jing and Wang, Bo and Yang, Ruichao},
  booktitle={Proceedings of the ACM Web Conference 2024},
  pages={2359--2370},
  year={2024}
}

@article{liu2025mind,
  title={MIND: A Multi-agent Framework for Zero-shot Harmful Meme Detection},
  author={Liu, Ziyan and Fan, Chunxiao and Lou, Haoran and Wu, Yuexin and Deng, Kaiwei},
  journal={arXiv preprint arXiv:2507.06908},
  year={2025}
}

@inproceedings{lin2025ask,
  title={Ask, acquire, understand: A multimodal agent-based framework for social abuse detection in memes},
  author={Lin, Xuanrui and Jia, Chao and Ji, Junhui and Han, Hui and Naseem, Usman},
  booktitle={Proceedings of the ACM on Web Conference 2025},
  pages={4734--4744},
  year={2025}
}

@article{achiam2023gpt,
  title={Gpt-4 technical report},
  author={Achiam, Josh and Adler, Steven and Agarwal, Sandhini and Ahmad, Lama and Akkaya, Ilge and Aleman, Florencia Leoni and Almeida, Diogo and Altenschmidt, Janko and Altman, Sam and Anadkat, Shyamal and others},
  journal={arXiv preprint arXiv:2303.08774},
  year={2023}
}

@inproceedings{ji2024capalign,
  title={Capalign: Improving cross modal alignment via informative captioning for harmful meme detection},
  author={Ji, Junhui and Lin, Xuanrui and Naseem, Usman},
  booktitle={Proceedings of the ACM Web Conference 2024},
  pages={4585--4594},
  year={2024}
}

@inproceedings{shah2024memeclip,
    title = "MemeCLIP: Leveraging CLIP Representations for Multimodal Meme Classification",
    author = "Shah, Siddhant Bikram  and
      Shiwakoti, Shuvam  and
      Chaudhary, Maheep  and
      Wang, Haohan",
    booktitle = "Proceedings of the 2024 Conference on Empirical Methods in Natural Language Processing",
    month = nov,
    year = "2024",
    address = "Miami, Florida, USA",
    publisher = "Association for Computational Linguistics",
    url = "https://aclanthology.org/2024.emnlp-main.959/",
    doi = "10.18653/v1/2024.emnlp-main.959",
    pages = "17320--17332",
}

@inproceedings{pramanick2021momenta,
  title={MOMENTA: A Multimodal Framework for Detecting Harmful Memes and Their Targets},
  author={Pramanick, Shraman and Sharma, Shivam and Dimitrov, Dimitar and Akhtar, Md Shad and Nakov, Preslav and Chakraborty, Tanmoy},
  booktitle={Findings of the Association for Computational Linguistics: EMNLP 2021},
  pages={4439--4455},
  year={2021}
}

@article{bai2023qwen,
  title={Qwen-vl: A frontier large vision-language model with versatile abilities},
  author={Bai, Jinze and Bai, Shuai and Yang, Shusheng and Wang, Shijie and Tan, Sinan and Wang, Peng and Lin, Junyang and Zhou, Chang and Zhou, Jingren},
  journal={arXiv preprint arXiv:2308.12966},
  volume={1},
  number={2},
  pages={3},
  year={2023}
}

@inproceedings{guo2020guoym,
  title={Guoym at SemEval-2020 task 8: Ensemble-based classification of visuo-lingual metaphor in memes},
  author={Guo, Yingmei and Huang, Jinfa and Dong, Yanlong and Xu, Mingxing},
  booktitle={Proceedings of the Fourteenth Workshop on Semantic Evaluation},
  pages={1120--1125},
  year={2020}
}

@article{liu2024sarcasm,
  title={Sarcasm driven by sentiment: A sentiment-aware hierarchical fusion network for multimodal sarcasm detection},
  author={Liu, Hao and Wei, Runguo and Tu, Geng and Lin, Jiali and Liu, Cheng and Jiang, Dazhi},
  journal={Information Fusion},
  volume={108},
  pages={102353},
  year={2024},
  publisher={Elsevier}
}

@inproceedings{hee2022explaining,
  title={On explaining multimodal hateful meme detection models},
  author={Hee, Ming Shan and Lee, Roy Ka-Wei and Chong, Wen-Haw},
  booktitle={Proceedings of the ACM web conference 2022},
  pages={3651--3655},
  year={2022}
}

@article{kingma2014adam,
  title={Adam: A method for stochastic optimization},
  author={Kingma, Diederik P},
  journal={arXiv preprint arXiv:1412.6980},
  year={2014}
}

@article{liu2026yes,
  title={Yes FLoReNce, I Will Do Better Next Time! Agentic Feedback Reasoning for Humorous Meme Detection},
  author={Liu, Olivia Shanhong and Ng, Pai Chet and Soh, De Wen and Plataniotis, Konstantinos N},
  journal={arXiv preprint arXiv:2601.07232},
  year={2026}
}

@article{rizwan2026see,
  title={See, Explain, and Intervene: A Few-Shot Multimodal Agent Framework for Hateful Meme Moderation},
  author={Rizwan, Naquee and Swain, Subhankar and Bhaskar, Paramananda and Aryan, Gagan and Khan, Shehryaar Shah and Mukherjee, Animesh},
  journal={arXiv preprint arXiv:2601.04692},
  year={2026}
}

@inproceedings{rizwan2025exploring,
  title={Exploring the Limits of Zero Shot Vision Language Models for Hate Meme Detection: The Vulnerabilities and their Interpretations},
  author={Rizwan, Naquee and Bhaskar, Paramananda and Das, Mithun and Majhi, Swadhin Satyaprakash and Saha, Punyajoy and Mukherjee, Animesh},
  booktitle={Proceedings of the International AAAI Conference on Web and Social Media},
  volume={19},
  pages={1669--1689},
  year={2025}
}

@inproceedings{hee2025contrastive,
  title={Contrastive Instruction Fine-Tuning Large Multimodal Model for Hateful Meme Classification},
  author={Hee, Ming Shan and Gao, Zihan and Wang, Yinglong and Chu, Xiangxiang and Lee, Roy Ka-Wei and Qin, Zengchang},
  booktitle={Proceedings of the International AAAI Conference on Web and Social Media},
  volume={19},
  pages={760--773},
  year={2025}
}

@inproceedings{thapa2025multimodal,
  title={A Multimodal Prompt-based Framework for Analyzing Code-Mixed and Low-Resource Memes},
  author={Thapa, Surendrabikram and Veeramani, Hariram and Hu, Liang and Zhang, Qi and Wang, Wei and Naseem, Usman},
  booktitle={Proceedings of the International AAAI Conference on Web and Social Media},
  volume={19},
  pages={1913--1923},
  year={2025}
}

@inproceedings{steffen2025more,
  title={More than Memes: A Multimodal Topic Modeling Approach to Conspiracy Theories on Telegram},
  author={Steffen, Elisabeth},
  booktitle={Proceedings of the International AAAI Conference on Web and Social Media},
  volume={19},
  pages={1831--1844},
  year={2025}
}

@inproceedings{crawford2021memes,
  title={Memes, radicalisation, and the promotion of violence on Chan sites},
  author={Crawford, Blyth and Keen, Florence and Suarez-Tangil, Guillermo},
  booktitle={Proceedings of the international AAAI conference on web and social media},
  volume={15},
  pages={982--991},
  year={2021}
}

@article{liu2023visual,
  title={Visual instruction tuning},
  author={Liu, Haotian and Li, Chunyuan and Wu, Qingyang and Lee, Yong Jae},
  journal={Advances in neural information processing systems},
  volume={36},
  pages={34892--34916},
  year={2023}
}

\section{Paper Checklist}
\begin{enumerate}

\item For most authors...
\begin{enumerate}
    \item  Would answering this research question advance science without violating social contracts, such as violating privacy norms, perpetuating unfair profiling, exacerbating the socio-economic divide, or implying disrespect to societies or cultures?
    \answerTODO{\textcolor{blue}{Yes, our work primarily focuses on utilizing VLMs to analyze and generate interpretations of humorous and hateful memes. While these generated interpretations may reflect social stereotypes, our goal is to enhance hateful meme detection systems and improve the understanding of such content.}}
  \item Do your main claims in the abstract and introduction accurately reflect the paper's contributions and scope?
    \answerTODO{\textcolor{blue}{Yes.}}
   \item Do you clarify how the proposed methodological approach is appropriate for the claims made? 
    \answerTODO{\textcolor{blue}{Yes.}}
   \item Do you clarify what are possible artifacts in the data used, given population-specific distributions?
    \answerTODO{\textcolor{blue}{Yes.}}
  \item Did you describe the limitations of your work?
    \answerTODO{\textcolor{blue}{Yes. You  may  find  them  under ``Limitations and Future Work" section.}}
  \item Did you discuss any potential negative societal impacts of your work?
    \answerTODO{\textcolor{blue}{Yes. You  may  find  them  under ``Ethics and Broader Impact" section.}}
      \item Did you discuss any potential misuse of your work?
    \answerTODO{\textcolor{blue}{Yes. You  may  find  them  under ``Ethics and Broader Impact" section.}}
    \item Did you describe steps taken to prevent or mitigate potential negative outcomes of the research, such as data and model documentation, data anonymization, responsible release, access control, and the reproducibility of findings?
    \answerTODO{\textcolor{gray}{N/A}}
  \item Have you read the ethics review guidelines and ensured that your paper conforms to them?
    \answerTODO{\textcolor{blue}{Yes.}}
\end{enumerate}

\item Additionally, if your study involves hypotheses testing...
\begin{enumerate}
  \item Did you clearly state the assumptions underlying all theoretical results?
    \answerTODO{\textcolor{gray}{N/A}}
  \item Have you provided justifications for all theoretical results?
    \answerTODO{\textcolor{gray}{N/A}}
  \item Did you discuss competing hypotheses or theories that might challenge or complement your theoretical results?
    \answerTODO{\textcolor{gray}{N/A}}
  \item Have you considered alternative mechanisms or explanations that might account for the same outcomes observed in your study?
    \answerTODO{\textcolor{gray}{N/A}}
  \item Did you address potential biases or limitations in your theoretical framework?
    \answerTODO{\textcolor{gray}{N/A}}
  \item Have you related your theoretical results to the existing literature in social science?
    \answerTODO{\textcolor{gray}{N/A}}
  \item Did you discuss the implications of your theoretical results for policy, practice, or further research in the social science domain?
    \answerTODO{\textcolor{gray}{N/A}}
\end{enumerate}

\item Additionally, if you are including theoretical proofs...
\begin{enumerate}
  \item Did you state the full set of assumptions of all theoretical results?
    \answerTODO{\textcolor{gray}{N/A}}
	\item Did you include complete proofs of all theoretical results?
    \answerTODO{\textcolor{gray}{N/A}}
\end{enumerate}

\item Additionally, if you ran machine learning experiments...
\begin{enumerate}
  \item Did you include the code, data, and instructions needed to reproduce the main experimental results (either in the supplemental material or as a URL)?
    \answerTODO{\textcolor{blue}{Yes. The Anonymous GitHub link can be found in the paper’s abstract.}}
  \item Did you specify all the training details (e.g., data splits, hyperparameters, how they were chosen)?
    \answerTODO{\textcolor{blue}{Yes. These information can be found under ``Implementation Details" section.}}

     \item Did you report error bars (e.g., with respect to the random seed after running experiments multiple times)?
    \answerTODO{\textcolor{blue}{No. The results reported in Tables 3, 4, and 5 are based on a single run using a fixed random seed (42) to ensure reproducibility.}}
	\item Did you include the total amount of compute and the type of resources used (e.g., type of GPUs, internal cluster, or cloud provider)?
    \answerTODO{\textcolor{blue}{Yes. These information can be found under ``Implementation Details" section.}}
     \item Do you justify how the proposed evaluation is sufficient and appropriate to the claims made? 
    \answerTODO{\textcolor{blue}{Yes. These information can be found under ``Experiments" section.}}
     \item Do you discuss what is ``the cost`` of misclassification and fault (in)tolerance?
    \answerTODO{\textcolor{gray}{N/A}}
  
\end{enumerate}

\item Additionally, if you are using existing assets (e.g., code, data, models) or curating/releasing new assets, \textbf{without compromising anonymity}...
\begin{enumerate}
  \item If your work uses existing assets, did you cite the creators?
    \answerTODO{\textcolor{blue}{Yes.}}
  \item Did you mention the license of the assets?
    \answerTODO{\textcolor{gray}{N/A}}
  \item Did you include any new assets in the supplemental material or as a URL?
    \answerTODO{\textcolor{gray}{N/A}}
  \item Did you discuss whether and how consent was obtained from people whose data you're using/curating?
    \answerTODO{\textcolor{gray}{N/A}}
  \item Did you discuss whether the data you are using/curating contains personally identifiable information or offensive content?
    \answerTODO{\textcolor{gray}{N/A}}
  \item If you are curating or releasing new datasets, did you discuss how you intend to make your datasets FAIR?
    \answerTODO{\textcolor{gray}{N/A}}
  \item If you are curating or releasing new datasets, did you create a Datasheet for the Dataset? 
    \answerTODO{\textcolor{gray}{N/A}}
\end{enumerate}

\item Additionally, if you used crowd sourcing or conducted research with human subjects, \textbf{without compromising anonymity}...
\begin{enumerate}
  \item Did you include the full text of instructions given to participants and screenshots?
    \answerTODO{\textcolor{teal}{No, we provided the annotators with instructions during meetings. All annotators are expert inthis field and no explicit instruction set was required}}
  \item Did you describe any potential participant risks, with mentions of Institutional Review Board (IRB) approvals?
    \answerTODO{\textcolor{gray}{N/A}}
  \item Did you include the estimated hourly wage paid to participants and the total amount spent on participant compensation?
    \answerTODO{\textcolor{teal}{No, all annotators were from authors’ institute.}}
   \item Did you discuss how data is stored, shared, and deidentified?
   \answerTODO{\textcolor{gray}{N/A}}
\end{enumerate}

\end{enumerate}

\appendix
\section{Appendix}
\subsection{Computing Resources}
We conducted our experiments on a high-performance computing server equipped with two AMD EPYC 7543 32‑Core Processors (64 cores per socket with simultaneous multithreading enabled, totaling 128 logical CPUs). The system has 514 GiB of DDR4 system memory, providing ample capacity for large‑scale data loading and preprocessing. The machine is outfitted with seven NVIDIA L40S GPUs, each offering 46,068 MiB of VRAM, enabling large‑batch multimodal training and inference.
The software environment includes Ubuntu 20.04, CUDA 12.5, and PyTorch 2.x with mixed precision training enabled via \texttt{torch.float16}.
The high-performance environment, particularly the 46,068 MiB VRAM of the L40S GPUs, allowed for the parallelized processing of the twelve perspective-conditioned reasoning traces during data preparation. This hardware setup ensures that the system can handle the high-throughput requirements of real-time social media moderation. The total training time for the entire MAR-12 framework on the Memotion dataset (approximately 5,600 memes) was completed in under 30 minutes, demonstrating that the framework is not only high-performing but also highly efficient to train.

For MAR‑12, we used CLIP ViT‑B/32 as the vision encoder, with image inputs resized to 224×224 pixels. The agent text embeddings were obtained using T5 (or CLIP text encoder in some variants) and projected into a 1024‑dimensional mapped space from the original 768‑dimensional embeddings. The fusion module included 1 mapping layer and 1 pre‑output layer, with dropout probabilities of 0.1, 0.4, and 0.2 applied at different stages. The classification head employed a cosine‑similarity‑based prototype classifier with a scale of 30 and margin ratio of 0.2.
Training was performed for 20 epochs using the AdamW optimizer with a learning rate of $1 \times 10^{-4}$, weight decay of $1 \times 10^{-4}$, and a batch size of 32. We used 4 data‑loading workers and fixed the random seed to 42 for reproducibility. Object‑level cues were extracted with a pretrained Faster R‑CNN (ResNet‑50‑FPN) detector (IoU threshold=0.5, score threshold=0.5). Object features were encoded with VGG‑16 (final classification layer removed), and attribute features were extracted using DistilBERT with uncased tokenization. All image features were normalized using ImageNet statistics.

\subsection{Training Configurations}
We applied a consistent training setup across all models to ensure comparability. Unless otherwise stated, We adopted a unified training strategy across all models:
\begin{itemize}
    \item \textbf{Optimizer:} AdamW
    \item \textbf{Learning Rate:} $1 \times 10^{-4}$
    \item \textbf{Weight Decay:} $1 \times 10^{-4}$
    \item \textbf{Batch Size:} 32
    \item \textbf{Epochs:} 5
    \item \textbf{Loss Function:} Cross-Entropy Loss for binary classification (2 output classes)
    \item \textbf{Seed:} 42 (for reproducibility)
\end{itemize}
All models were trained on \texttt{image\_size=224} inputs (when applicable), with \texttt{num\_workers=4} used for data loading.
Cross-entropy loss was used as the objective function for all binary classification tasks. Mixed precision training was enabled throughout to improve computational efficiency and memory utilization.
The rapid training time ($<$ 30 minutes for 20 epochs) is a direct result of MAR-12's lightweight design. Unlike competitive Large Multimodal Model (LMM) baselines like LoReHM (34.00B trainable parameters), MAR-12 utilizes only 1.31M trainable parameters.
Because the heavy VLM (Qwen-VL) and encoders (CLIP/T5) are kept frozen, the training process focuses solely on the role-aware attention and the prototype-based classification head. This architectural choice significantly reduces the computational overhead and carbon footprint of model development compared to end-to-end fine-tuning.

\subsection{Inference Complexity and Scalability}
Inference Complexity and Scalability. At inference time, MAR-12 processes a single meme in an average of 3.5 seconds. This duration accounts for the end-to-end pipeline: (1) twelve parallelized VLM forward passes to generate reasoning traces, (2) the soft-gated attention aggregation, (3) the prototype-based classification, and (4) the final LLM-based explanation synthesis.
While the 12 VLM passes introduce a higher cost than a single black-box classification, this cost is offset by the quality and interpretability of the results. To improve scalability, our implementation utilizes batch processing for the 12 perspective prompts, reducing redundant vision-encoding passes. Compared to iterative multi-agent debate frameworks which can take upwards of 10–15 seconds per meme, MAR-12’s 3.5-second latency provides a superior balance between deep reasoning and the response times required for large-scale content moderation.

\subsection{Evaluation Metrics}
We evaluate model performance using Accuracy (ACC), Area Under the Receiver Operating Characteristic Curve (AUC), and F1‑score. Accuracy measures the proportion of correctly classified memes and serves as a straightforward indicator of overall predictive performance. However, given that humour and hate meme detection tasks can exhibit class imbalance (Table 2) and involve ambiguous, borderline cases, we complement accuracy with additional metrics.
AUC evaluates the model’s ability to rank positive and negative samples correctly across varying classification thresholds, making it less sensitive to class distribution and more informative for imbalanced settings. F1‑score, the harmonic mean of precision and recall, is particularly important in this context: a model that predicts most memes as “humour” or “hate” may achieve high accuracy but poor balance between false positives and false negatives. 

We followed the official train/validation/test splits for the PrideMM and Memotion datasets. During evaluation, we computed accuracy, AUC, and F1‑score using the official labels, and additionally saved full classification reports, confusion matrices, and per‑meme prediction logs.
All results reported in the paper are based on a single run using a fixed random seed for reproducibility.

\subsection{Baseline Model-Specific Details}
\paragraph{MemeCLIP \citep{shah2024memeclip}}
This model builds upon the CLIP ViT-L/14 encoder, with frozen parameters. Image and text features are independently projected using linear mapping layers, followed by adapter modules for task-specific refinement. The projected and adapted embeddings are normalized and fused via element-wise multiplication. The final representation is passed through an optional pre-output MLP and a cosine-based classifier. Key architectural parameters include:
\begin{itemize}
    \item \texttt{unmapped\_dim=768}, \texttt{map\_dim=1024}
    \item \texttt{drop\_probs=[0.1, 0.4, 0.2]}
    \item \texttt{ratio=0.2} for adapter residual weighting
    \item Cosine classifier with scaling factor \texttt{scale=30}
\end{itemize}

\paragraph{MOMENTA \citep{pramanick2021momenta}}
MOMENTA leverages both object-level and attribute-level visual signals fused with CLIP image and text features. Intra-modal fusion is performed via a custom cross-modal attention mechanism, which aligns CLIP embeddings with extracted features (e.g., from object detection or attribute graphs). The two streams are then aggregated via a self-attention fusion block to form a unified multimodal representation. The final prediction is produced by a linear classifier. This model captures both intra- and inter-modal dependencies with fine-grained attention.

\paragraph{PromptHate \citep{cao2022prompting}}
We use a RoBERTa-based prompting model for binary classification. The model architecture is based on \texttt{roberta-large}, trained in a masked language modeling setup. Task-specific verbalizers are used to map label tokens to the logits extracted at the masked position. This approach aligns with prompt-based learning paradigms where no classifier head is introduced — the prediction is directly computed based on the masked token output distribution.

\paragraph{LoReHM \citep{huang2024towards}}
We utilize  LLaVA34B  as the LMM agent from the
the open-source perspectives.
Specifically, we implement the “llava-v1.6-34b”
for LLaVA-34B. The frozen pretrained vision and text Transformer encoders are implemented as CLIP  with
the specific version “ViT-L/14@336px”.

\paragraph{MAR-12 (with Role-aware Attention-based Aggregator).}
Our proposed MAR-12 model integrates vision-language grounding with structured reasoning via twelve role-specific agents. Each agent produces an embedding projected into a shared feature space using agent-specific MLPs. Attention is computed across agent embeddings, yielding a weighted representation that captures the most salient reasoning cues.

The visual input is encoded using CLIP ViT-L/14, and both the agent and image embeddings are passed through shared mapping and adapter layers (similar to MemeCLIP). The final representation is computed by element-wise multiplication and classified using a cosine-based classifier. This architecture supports interpretability and extensibility via role-aware agent prompting.

A fixed random seed (42) was used across all experiments to ensure reproducibility.

Table~\ref{tab:model_summary} lists the parameter count for all models.


\section{Appendix: Prompting Strategy, Psychological Grounding, and Explainer Prompt}
\label{app:prompting}
\subsection{Design Goals and Rationale}
Our prompting strategy is built to (i) \textbf{separate roles} (facts $\rightarrow$ interpretation), (ii) \textbf{cover complementary mechanisms} of humour and harm with orthogonal lenses, and (iii) \textbf{constrain outputs} so that the attention module can weight agent evidence and the explainer can \emph{faithfully} summarize that attention-weighted reasoning.

\paragraph{Role separation.}
We elicit \emph{foundational facts} (scene description, OCR) with report-only prompts that explicitly prohibit interpretation, then invite \emph{interpretive} agents (humour and safety/intent) to reason on top of those facts. This reduces leakage and makes attention assignments diagnostic.

\paragraph{Mechanism coverage.}
Humour in memes is multifactorial. We therefore cover visual/textual incongruity, affective contrast, cultural schema, absurdity, wordplay, timing/punchline structure, and image--text alignment. For safety, we cover harmful impact (hatefulness) and authorial motive (intent), recognizing that harmful memes often co-opt humour mechanics.

\paragraph{Output constraints.}
Prompts use imperative, scope-limited language (e.g., ``avoid interpretation'', ``return text only'') so agent outputs are short, typed rationales that are easy to weight (by attention) and cite (by the explainer). This supports process-level faithfulness rather than label-chasing.

\subsection{Agent Prompts (12 Roles)}
\begin{table*}[t]
\centering
\small
\resizebox{\linewidth}{!}{
\begin{tabular}{|p{2cm}|p{4cm}|p{12cm}|}
\hline
\textbf{Role} & \textbf{Agent Name} & \textbf{Prompt} \\
\hline
\multirow{2}{2cm}{Foundational Analysis}
& General Description Agent & \textit{Provide a detailed description of the image. What objects, characters, facial expressions, and scene elements are visible? Avoid interpretation, just describe what is shown.} \\ \cline{2-3}
& OCR \& Text Extraction Agent & \textit{Extract all text visible in the image, including captions, embedded text, and any signs or symbols. Return the extracted text only.} \\
\hline
\multirow{8}{2cm}{Humour-Oriented Reasoning}

& Visual Irony Agent & \textit{Look at the visual content of the image. Is there any contradiction, mismatch, or unexpected element that creates irony or humour? Describe how the image visually subverts expectations.} \\  \cline{2-3}
& Textual Irony Agent & \textit{Read the text in the image. Is there irony, sarcasm, or a contradiction between what is said and what is implied? Explain whether this contributes to humour.} \\ \cline{2-3}
& Emotion Contrast Agent & \textit{Analyze the emotional tone of the image (e.g., facial expressions, setting) and compare it with the tone of the text. Is there a humorous contrast or mismatch between the two?} \\ \cline{2-3}

& Cultural Reference Agent & \textit{Does the meme refer to any cultural, social, political, or internet-related topic that contributes to its humour? Identify and explain any references that are key to understanding the joke.} \\ \cline{2-3}

& Absurdity Check Agent & \textit{Does the meme contain elements that are absurd, exaggerated, or nonsensical in a way that is intended to be funny? If yes, describe them and how they contribute to humour.} \\ \cline{2-3}

& Pun \& Wordplay Check Agent & \textit{Check if the meme contains puns, double meanings, rhymes, or wordplay. If so, explain how the text plays with language to create humour.} \\ \cline{2-3}

& Punchline Check Agent & \textit{Does the meme follow a setup and punchline structure? Identify the setup and the punchline, and explain how the timing or surprise of the punchline contributes to humour.} \\ \cline{2-3}

& Image-Text Alignment Agent & \textit{Evaluate the relationship between the image and the text. Do they work together to create humour, or do they contradict each other? Explain how their alignment or misalignment affects the meme’s humour.} \\ \cline{2-3}
\hline

\multirow{2}{2cm}{Safety \& Intent Evaluation}
& Hatefulness Detection Agent & \textit{Analyze the meme to determine if it includes offensive, derogatory, or hateful content. Could the meme harm or marginalize any group of people, even if it’s intended to be funny? Provide reasoning.} \\ \cline{2-3}
& Intent Interpretation Agent & \textit{Based on the image and text, infer what the creator intended. Was it meant to entertain, criticize, mock, or provoke? Is the humour light-hearted or could it be seen as targeted or aggressive? Explain your interpretation.} \\ \cline{2-3}
\hline
\end{tabular}}
\caption{Overview of the 12 specialized agents used in MAR-12, including their roles and associated prompts.}
\label{tab:agent-prompts-appendix}
\end{table*}

\subsection{Explainer Prompt and Interface to Attention}
The explainer receives three inputs: (1) the final prediction $\hat{y}$, (2) the set of agent responses $\{r_i\}_{i=1}^{12}$, and (3) attention weights $\{\alpha_i\}_{i=1}^{12}$ from the soft aggregation module. The explainer is instructed to summarize \emph{why} the system decided on $\hat{y}$ by following the attention-weighted agent evidence.

\paragraph{Explainer Prompt.}
\begin{quote}
\textit{``Given the prediction: [LABEL], and the following agent insights \underline{weighted by relevance} ($\alpha_i$), summarize the reasoning behind the classification. Cite the most influential agents (higher $\alpha$) and explain how their evidence supports the decision. If there are conflicting signals from lower-weighted agents, briefly acknowledge them.''}
\end{quote}

\subsection{Psychological Grounding of Prompt Families}
\textbf{Incongruity \& resolution} (visual/textual irony; image--text alignment) motivate prompts that explicitly test for mismatches and their resolution.  
\textbf{Affective contrast} (emotion vs.\ content) motivates prompts that ask for tonal mismatches.  
\textbf{Schema/culture dependence} motivates prompts for cultural references to supply background needed for comprehension.  
\textbf{Absurdist/nonsense humour} motivates prompts to flag exaggeration/surrealism.  
\textbf{Linguistic ambiguity} motivates puns/wordplay prompts.  
\textbf{Timing/surprise (setup--punchline)} motivates explicit structure prompts.  
\textbf{Superiority/benign violation/social identity harms} motivate hatefulness prompts (impact on groups).  
\textbf{Theory of mind/Gricean pragmatics} motivates intent prompts (speaker meaning vs.\ literal form).

\subsection{Why This Prompt Design Supports Faithfulness and Novelty}
\begin{itemize}
\item \textbf{Process alignment by construction.} Agents yield typed, local rationales; the attention head learns a sparse mixture over these; the explainer is instructed to mirror that mixture. This produces explanations that track the system’s internal evidence flow (role-aware soft attention) rather than post-hoc label rationalization.
\item \textbf{Separation of facts and interpretation.} Report-only foundational prompts reduce contamination and make later interpretive weights more meaningful.
\item \textbf{Dual-use lenses for humour vs.\ harm.} The same mechanisms that make content funny can also make it harmful; prompting both humour and safety over the \emph{same} lenses lets the model explain how those mechanisms were deployed (benign vs.\ targeted).
\item \textbf{Extensibility.} New lenses can be added as new agents without disrupting others; the attention layer simply learns when they matter, and the explainer inherits them in explanation.
\end{itemize}

\subsection{Implementation Notes (for Reproducibility)}
\begin{itemize}
\item Prompts use imperative phrasing (``identify whether'', ``explain how'') and scope guards (``avoid interpretation'', ``return text only'') to standardize outputs.
\item Agent responses should be concise (2--5 sentences or bullet points), focusing on evidence, not conclusions; this improves attention calibration and explainer citability.
\item The explainer should be passed $(r_i, \alpha_i)$ pairs in descending $\alpha_i$ order to bias summaries toward most influential evidence while still acknowledging conflicts.
\end{itemize}


\section{ Error Analysis}
\begin{figure}[h]
    \centering
    \includegraphics[width=0.45\textwidth]{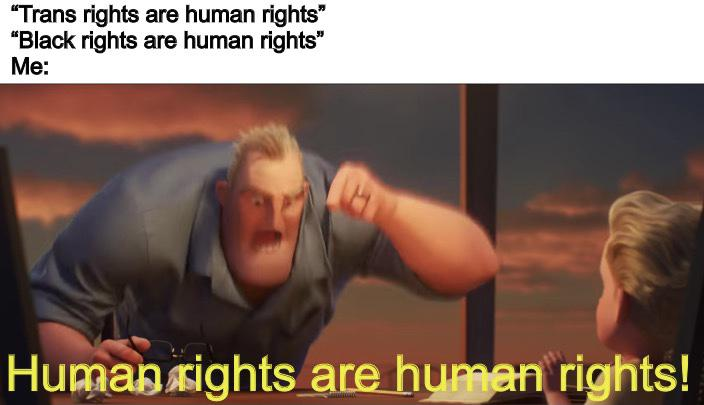}
    \caption{Label: No Humour, No Hate; Prediction: \textcolor{red}{Humour, Hate} }
    \label{fig:4074}
\end{figure}

\begin{itemize}
    \item The image shows two animated characters in an office setting. One character, a large man with a stern expression, is leaning forward with his fist raised, seemingly in a confrontational stance. The other character, a smaller person with blonde hair, is seated across from him, looking up at the larger character. The background features a window showing a sunset sky. Text at the top reads "Trans rights are human rights" and "Black rights are human rights," while text at the bottom says "Human rights are human rights!"

    \item OCR \& Text Extraction Agent: "Trans rights are human rights" "Black rights are human rights" Me:
"Human rights are human rights!"

    \item Visual Irony Agent: The image humorously subverts expectations by presenting a character who appears to be advocating for human rights in a dramatic manner, yet the text at the bottom states "Human rights are human rights!" This repetition and the character's exaggerated expression create an ironic contrast, suggesting a playful dismissal of the serious topic through over-the-top enthusiasm.
    
    \item Textual Irony Agent: The text implies that all human rights are inherently equal, which contrasts with the implication of prioritizing certain groups over others. This creates a humorous effect by highlighting the absurdity of the statement through its own logical inconsistency.
    
    \item Emotion Contrast Agent: The image conveys a sense of frustration and intensity through the character's exaggerated facial expression and body language, while the text presents a straightforward statement about human rights. This creates a humorous contrast, as the dramatic reaction seems disproportionate to the simple assertion made in the text.
     
     \item Cultural Reference Agent: The meme humorously addresses the intersection of various human rights movements by juxtaposing the phrases "Trans rights are human rights" and "Black rights are human rights," suggesting that these are part of a broader, unified concept of human rights. The character's exaggerated reaction emphasizes the importance of this unity, making the joke resonate with those who support the idea that all human rights are interconnected and essential.
     
     \item Absurdity Check Agent: Yes, the meme uses exaggerated body language and a dramatic pose from an animated character to humorously emphasize the point that all human rights are equal, making it absurdly clear and funny.
     
    \item Pun \& Wordplay Check Agent: The meme uses wordplay by repeating "human rights" in the text, creating a humorous echo that emphasizes the importance of human rights for all groups, including trans and black individuals.
    
    \item Punchline Check Agent: Yes, the meme follows a setup and punchline structure. The setup is the statement "Trans rights are human rights" and "Black rights are human rights," leading to the punchline "Human rights are human rights!" The humor comes from the unexpected repetition of the phrase "human rights," which emphasizes the universality of human rights and creates a comedic effect by highlighting the redundancy.
    
    \item Image-Text Alignment Agent: The image and text work together to create humor by using the exaggerated reaction of the character to emphasize the point that "Human rights are human rights!" This misalignment between the serious statement about human rights and the over-the-top reaction adds to the comedic effect, as it contrasts the expected calmness with the unexpected intensity, suggesting a humorous disbelief or frustration at the idea being presented
    
    \item Hatefulness Detection Agent: The meme does not include offensive, derogatory, or hateful content. It humorously emphasizes the universal nature of human rights by highlighting that both trans rights and black rights are part of the broader concept of human rights, without targeting any specific group negatively.
    
    \item Intent Interpretation Agent: The creator likely intended to entertain with a humorous take on the concept of human rights, using exaggerated body language for comedic effect. The meme could also be seen as mildly provocative by highlighting the importance of recognizing all human rights equally.  
\end{itemize}

In this instance, the meme was misclassified by the MAR-12 as both humorous and hateful, despite the agent responses suggesting it does not contain hateful content. Specifically, the Hatefulness Detection Agent determined that the meme does not include offensive or derogatory elements, but instead humorously underscores the universality of human rights by affirming that both trans rights and Black rights are integral to human rights, without negatively targeting any group. Meanwhile, the Intent Interpretation Agent inferred that the creator’s aim was to entertain by exaggerating body language for comedic effect, perhaps also prompting reflection on the importance of recognizing all human rights equally.

The misclassification likely stems from the subjectivity inherent in human annotations, particularly in content addressing sensitive topics where interpretations of humor and offense can vary greatly among annotators. While the meme’s exaggerated presentation might have led some annotators to perceive it as mocking or trivializing serious issues, the agents interpreted it as a humorous yet earnest commentary on equality. This divergence underscores the challenge of aligning model outputs with subjective human judgments in nuanced, socially charged content. Addressing this challenge will require future work on better modeling subjectivity in humour and hate detection, including training with datasets that capture annotator disagreement, and exploring uncertainty-aware classification frameworks that can reflect ambiguous or borderline cases.

\begin{figure}[h]
    \centering
    \includegraphics[width=0.30\textwidth]{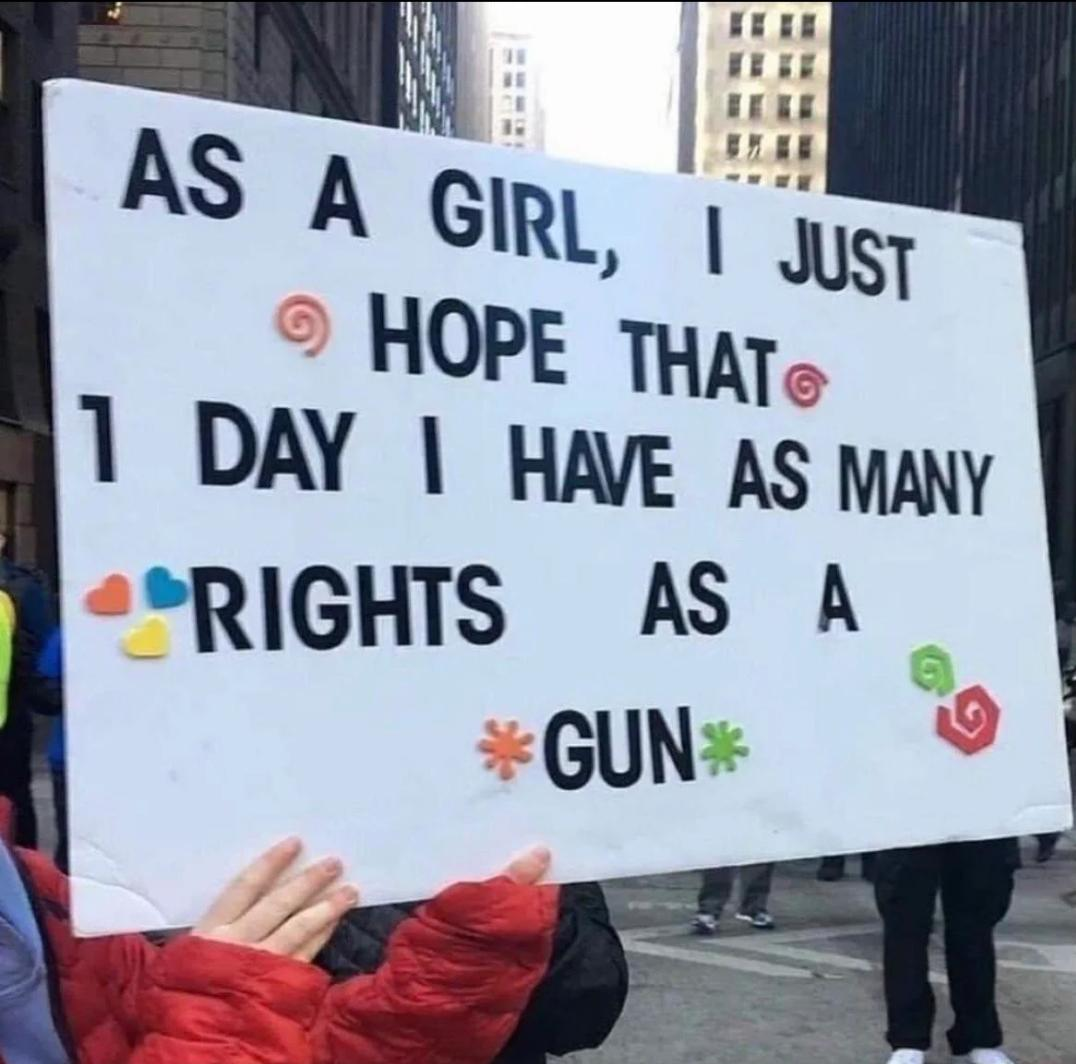}
    \caption{Label: Humour, Hate; Prediction: Humour \textcolor{red}{No Hate}}
    \label{fig:3893}
\end{figure}


\begin{itemize}
    \item General Description Agent: The image shows a protest sign held by someone wearing a red jacket. The sign reads: "AS A GIRL, I JUST HOPE THAT 1 DAY I HAVE AS MANY RIGHTS AS A GUN." The text is written in black letters on a white background, with decorative elements like hearts and stars around the words. The background includes buildings and other people, suggesting an urban setting during a public demonstration.

    \item OCR \& Text Extraction Agent: "AS A GIRL, I JUST HOPE THAT 1 DAY I HAVE AS MANY RIGHTS AS A GUN"

    \item Visual Irony Agent: The sign humorously contrasts the serious topic of gender equality with a playful, nonsensical comparison to gun rights, creating an ironic and unexpected juxtaposition that subverts expectations by using a lighthearted tone to address a serious issue.
    
    \item Textual Irony Agent: The sign expresses a serious hope for equality in rights, but the use of "gun" as a metaphor for rights creates an ironic and humorous contrast, playing on the absurdity of equating rights with a dangerous object.
    
    \item Emotion Contrast Agent: The image conveys a serious and urgent tone, likely from a protest or demonstration, while the text on the sign uses a sarcastic and ironic tone, suggesting a humorous contrast by comparing rights to gun ownership, which is typically associated with power and control rather than equality.
     
     \item Cultural Reference Agent: The meme humorously critiques gender inequality by playfully suggesting that women should have as many rights as guns, which is a hyperbolic exaggeration meant to highlight the absurdity of the situation. The use of colorful stickers and playful fonts adds to the comedic effect by contrasting the serious issue with a lighthearted presentation.
     
     \item Absurdity Check Agent: Yes, the meme uses absurdity and exaggeration by comparing the rights of a girl to those of a gun, which is nonsensical and intended to be humorous by highlighting the irony and absurdity of the situation.
     
    \item Pun \& Wordplay Check Agent: The meme uses a pun by playfully juxtaposing ""rights"" with ""gun,"" creating a humorous and ironic statement about gender equality. The use of asterisks around ""gun"" adds to the absurdity, emphasizing the unexpected nature of the comparison.
    
    \item Punchline Check Agent: The meme follows a setup and punchline structure where the setup is the hopeful statement about having equal rights as a girl, and the punchline is the unexpected and ironic twist of equating rights with a gun. The humor comes from the stark contrast between the innocent hope for equality and the absurdity of suggesting that guns are a measure of rights, which subverts expectations and creates a humorous effect through irony.
    
    \item Image-Text Alignment Agent: The image and text work together to create humor by using a provocative and absurd juxtaposition. The sign's message about gender equality is paired with the word ""gun,"" which is typically associated with power and rights, but here it is used in a way that subverts expectations, making the statement both shocking and humorous.
    
    \item Hatefulness Detection Agent: The meme uses a provocative and insensitive metaphor comparing gender rights to gun ownership, which could be interpreted as a harmful or derogatory statement towards women's rights. It may marginalize women by suggesting that their rights are equivalent to those of a weapon, potentially perpetuating stereotypes about gender equality.
    
    \item Intent Interpretation Agent: The creator likely intended to provoke thought and highlight the absurdity of gender inequality by using a provocative statement that contrasts the desire for equal rights with the metaphorical "right" to bear arms. The humor is not light-hearted but rather targeted at drawing attention to serious issues through a bold and unconventional approach.   
\end{itemize}

The conflicting perspectives in the safety agents' responses highlight a core challenge in humor detection: the subjective nature of humor itself, especially when intertwined with sensitive social topics. While human annotators may have perceived the meme as humorous, the agents interpreted the content as potentially harmful and therefore did not classify it as light-hearted humour. This divergence underscores how subjectivity in humour perception, both among human annotators and within the model’s reasoning process, can lead to classification disagreements, particularly in cases involving controversial or polarizing subjects.

\begin{figure}[h]
    \centering
    \includegraphics[width=0.30\textwidth]{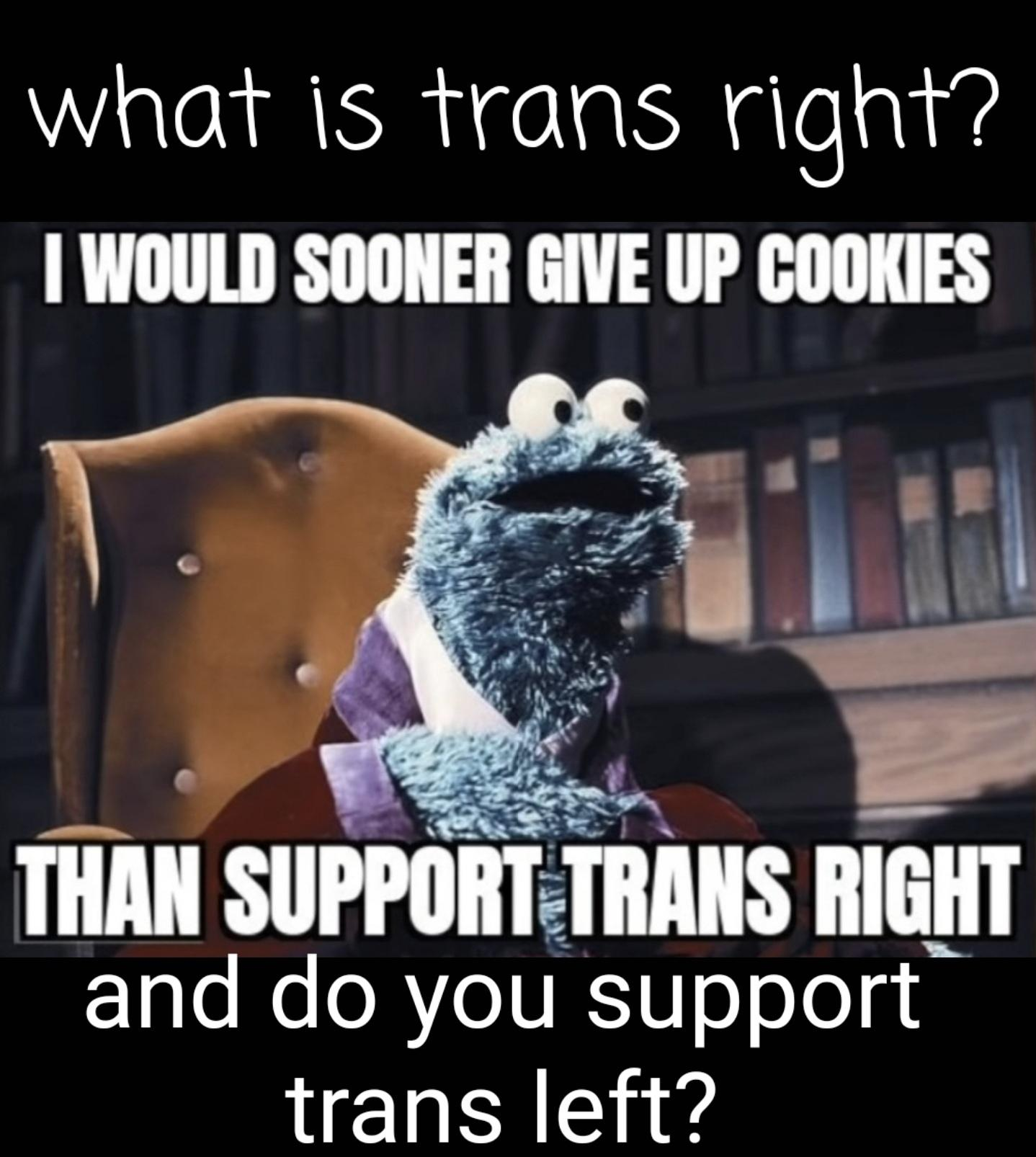}
    \caption{Label: Humour, Hate; Prediction: Humour,\textcolor{red}{No Hate}}
    \label{fig:4099}
\end{figure}


\begin{itemize}
    \item General Description Agent: The image features Cookie Monster from Sesame Street sitting in a chair with a bookshelf in the background. He has a surprised expression on his face. Text overlaid on the image reads: "what is trans right?"

    \item OCR \& Text Extraction Agent: "what is trans right? I WOULD SOONER GIVE UP COOKIES THAN SUPPORT TRANS RIGHT and do you support trans left?"

    \item Visual Irony Agent: The image humorously subverts expectations by presenting Cookie Monster as a character who would give up cookies rather than support trans rights, creating an ironic contrast between his typically cookie-loving persona and his unexpected stance on the issue.
    
    \item Textual Irony Agent: The sign expresses a serious hope for equality in rights, but the use of "gun" as a metaphor for rights creates an ironic and humorous contrast, playing on the absurdity of equating rights with a dangerous object.
    
    \item Emotion Contrast Agent: The image features Cookie Monster with a neutral expression, seated in a library-like setting, which contrasts humorously with the text that presents a serious and potentially controversial topic about trans rights. The juxtaposition creates a comedic effect by highlighting the absurdity of the statement.
     
     \item Cultural Reference Agent: The meme humorously references the complex and often contentious issue of trans rights, using a well-known character's exaggerated reluctance to support trans rights as a punchline. The joke plays on the absurdity of prioritizing cookies over supporting a marginalized group, highlighting the absurdity of some people's stance on the matter.
     
     \item Absurdity Check Agent: Yes, the meme uses absurd exaggeration by suggesting someone would give up cookies rather than support trans rights, which is meant to be humorous by highlighting the extreme nature of the statement. The juxtaposition of this extreme position with the question about supporting trans rights on the left adds to the comedic effect.
     
    \item Pun \& Wordplay Check Agent: The meme uses wordplay by playfully juxtaposing "trans right" and "trans left," creating a humorous contrast that relies on the audience's understanding of the term "trans" in both political contexts and its literal meaning. The phrase "I would sooner give up cookies than support trans right" is a clever pun, as it humorously suggests a strong preference for cookies over supporting transgender rights, playing on the idea of giving something up for a cause.
    
    \item Punchline Check Agent: The meme follows a setup and punchline structure where the setup asks about trans rights, leading to the punchline that humorously suggests giving up cookies rather than supporting trans rights, which is unexpected and absurd, contributing to its comedic effect.
    
    \item Image-Text Alignment Agent: The image of Cookie Monster, known for his love of cookies, paired with the text that humorously suggests he would give up cookies rather than support trans rights, creates a playful contradiction that adds to the humor by juxtaposing his character trait with the serious topic of trans rights. The alignment of the text with the image's absurdity enhances the meme's comedic effect.
    
    \item Hatefulness Detection Agent: The meme uses a character known for its controversial statements about trans rights, which can be seen as offensive and potentially harmful to the trans community by promoting a harmful narrative that equates support for trans rights with a lack of support for cookies. This could marginalize and alienate individuals who identify as trans or support their rights.
    
    \item Intent Interpretation Agent: The creator likely intended to provoke thought or criticism about the topic of trans rights, using a humorous yet potentially insensitive approach that might be seen as aggressive by some viewers.   
\end{itemize}

The Hatefulness Detection Agent identified that the meme features a character known for controversial statements about trans rights, framing the meme as potentially offensive and harmful by perpetuating a narrative that could marginalize or alienate trans individuals and their allies. This agent clearly signaled the presence of hateful or harmful content. However, the Intent Interpretation Agent suggested that the creator’s goal was to provoke thought or criticism around trans rights using humor, albeit in a manner that could be perceived as insensitive or aggressive by some audiences. This interpretation leans toward reading the meme as provocative satire rather than explicit hate.

\begin{figure}[h]
    \centering
    \includegraphics[width=0.30\textwidth]{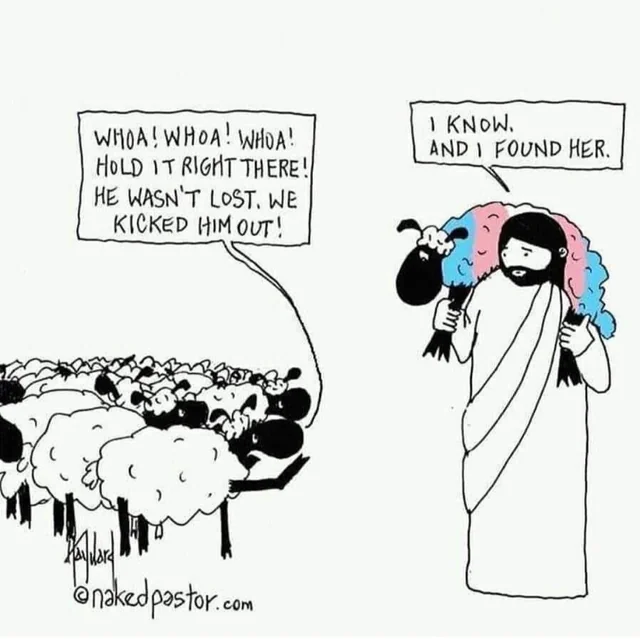}
    \caption{Label: No Humour, Hate; Prediction: \textcolor{red}{Humour, No Hate}}
    \label{fig:4095}
\end{figure}


\begin{itemize}
    \item General Description Agent:  The image is a cartoon featuring a shepherd holding a sheep with a rainbow-colored head. The shepherd says, ""I know. And I found her. On the left side, a flock of sheep is gathered around another figure, who is saying, ""Whoa! Whoa! Whoa! Hold it right there! He wasn't lost, we kicked him out!"" 

    \item OCR \& Text Extraction Agent: "WHOA! WHOA! WHOA! HOLD IT RIGHT THERE! HE WASN'T LOST, WE KICKED HIM OUT!
    I KNOW. AND I FOUND HER.

    \item Visual Irony Agent: The image humorously subverts expectations by depicting Jesus holding a sheep with a rainbow-colored head, which is an unexpected and whimsical twist on the traditional imagery of Jesus as the shepherd. The dialogue further adds to the irony, as it suggests that Jesus found someone who was not lost but was instead kicked out, creating a playful and absurd scenario.
    \item Textual Irony Agent: The text implies that Jesus found someone who was not lost but was kicked out, which creates a humorous contradiction with the typical narrative of Jesus finding lost sheep. This irony adds to the humor by subverting expectations.
    
    \item Emotion Contrast Agent: The image conveys a lighthearted and somewhat ironic tone through its simple line art and the unexpected twist of Jesus holding a sheep, which contrasts humorously with the serious implication of "kicking someone out."
     
     \item Cultural Reference Agent: The text, while straightforward, adds to this contrast by emphasizing the absurdity of the situation, creating a comedic effect.
     
     \item Absurdity Check Agent: The meme humorously plays on the idea of "kicking someone out" by referencing a common phrase used when someone is lost, suggesting that the person was never lost in the first place but was intentionally excluded. The use of Jesus holding a sheep with a rainbow-colored head adds a whimsical twist, enhancing the comedic effect.
     
    \item Pun \& Wordplay Check Agent: Yes, the meme contains absurd and exaggerated elements intended to be humorous. The depiction of Jesus holding a sheep with a beard and a speech bubble saying "I know. And I found her" alongside the flock's reaction of disbelief and concern creates a comical contrast between the expected and the unexpected. The meme uses wordplay by playfully misinterpreting "kicked him out" as "kicked him out" of a relationship, creating a humorous twist on the biblical story of Jesus finding the lost sheep.
    
    \item Punchline Check Agent: Yes, the meme follows a setup and punchline structure. The setup is the flock of sheep calling out to Jesus, thinking he's found someone lost, while the punchline is Jesus saying, "I know. And I found her." The humor comes from the unexpected twist that Jesus is referring to a female companion, which is a surprising and playful take on the traditional narrative of Jesus finding the lost sheep.
    
    \item Image-Text Alignment Agent: The image and text work together to create humor by presenting a humorous contradiction. The text suggests that someone was kicked out of the group, while the image depicts Jesus holding a sheep with a rainbow-colored headband, implying he found someone who wasn't lost but rather part of the group. This misalignment of expectations adds to the comedic effect.
    
    \item Hatefulness Detection Agent: The meme does not include offensive, derogatory, or hateful content. However, it could potentially be seen as insensitive by some viewers due to its reference to transgender individuals, which might be considered harmful or marginalizing to those who identify with the LGBTQ+ community.
    
    \item Intent Interpretation Agent: The creator likely intended to entertain with a humorous take on a biblical story, using irony and exaggeration to create a comedic effect. The use of "kicked him out" instead of "lost" adds a layer of absurdity that is meant to be lighthearted rather than aggressive or critical.   
\end{itemize}

The misclassification appears to stem from the dominant influence of the intent interpretation agent’s output, which emphasized the meme’s humorous intent and overshadowed the caution flagged by the hatefulness detection agent. This case illustrates a key limitation of the current attention-weighting mechanism, where insufficiently calibrated weights can cause certain agent perspectives to disproportionately sway the final decision. It underscores the importance of carefully balancing the contributions of each agent to ensure that signals indicating potential harm or offensiveness are not inadvertently diluted by agents emphasizing benign interpretations.


\section{Interpreting Agent-Level Attention and Its Implications}
\label{app:attn}

This appendix analyzes the learned attention over the twelve role-specialized agents and demonstrates how the model allocates evidential focus differently for humour versus hate detection. The results substantiate two central claims of our work: (i) the multi-agent, soft-attention architecture yields task-sensitive and human-interpretable evidence integration, and (ii) humour and hate are not mutually exclusive; rather, they rely on distinct but complementary evidential pathways that our framework can disentangle and recombine.

\begin{figure}[t]
  \centering
  \includegraphics[width=\linewidth]{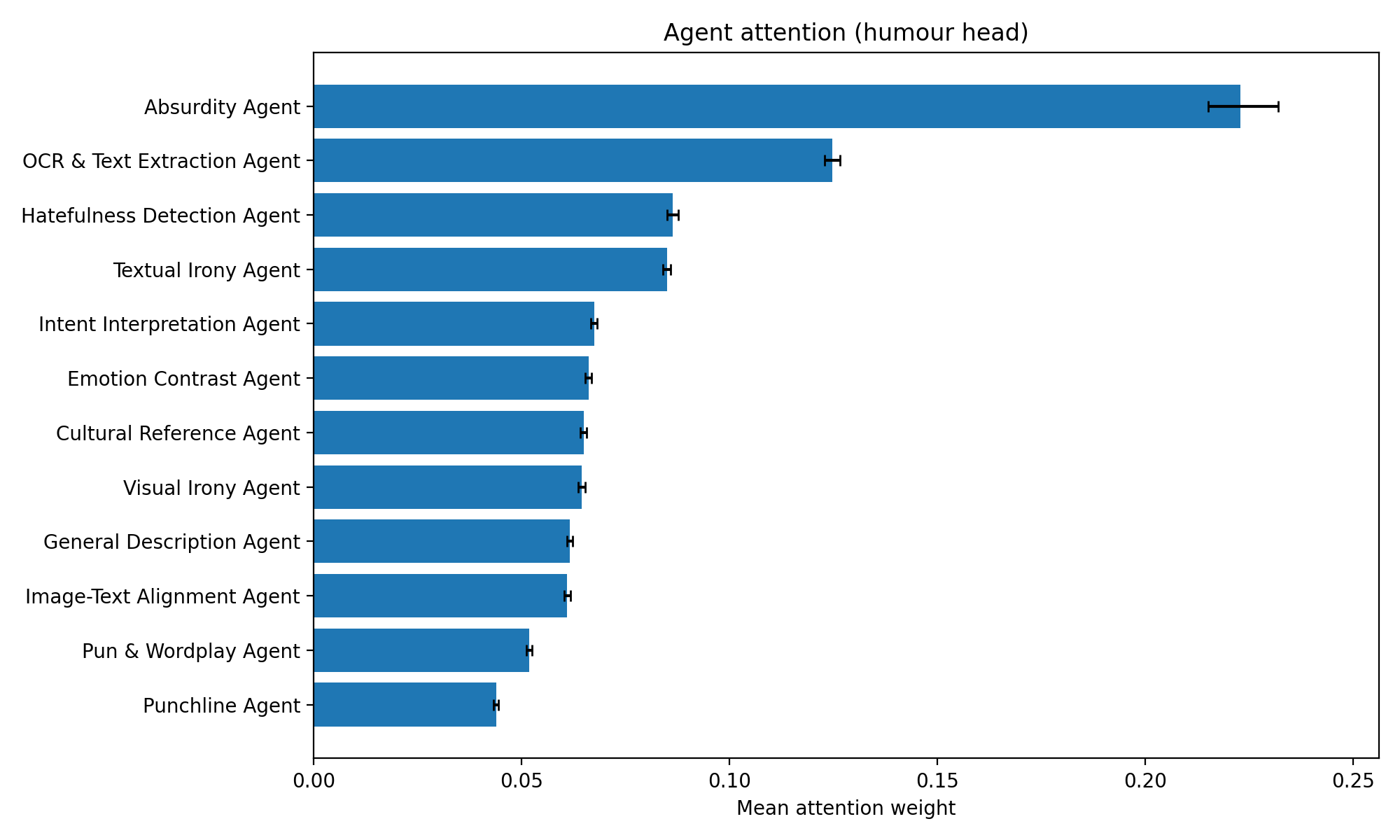}
  \caption{Ranked mean attention (with 95\% CI) over agents for the humour head. The model emphasizes \emph{Absurdity} and \emph{OCR \& Text Extraction}, followed by irony-, contrast-, and reference-focused agents.}
  \label{fig:ranked-humour}
\end{figure}

\begin{figure}[t]
  \centering
  \includegraphics[width=\linewidth]{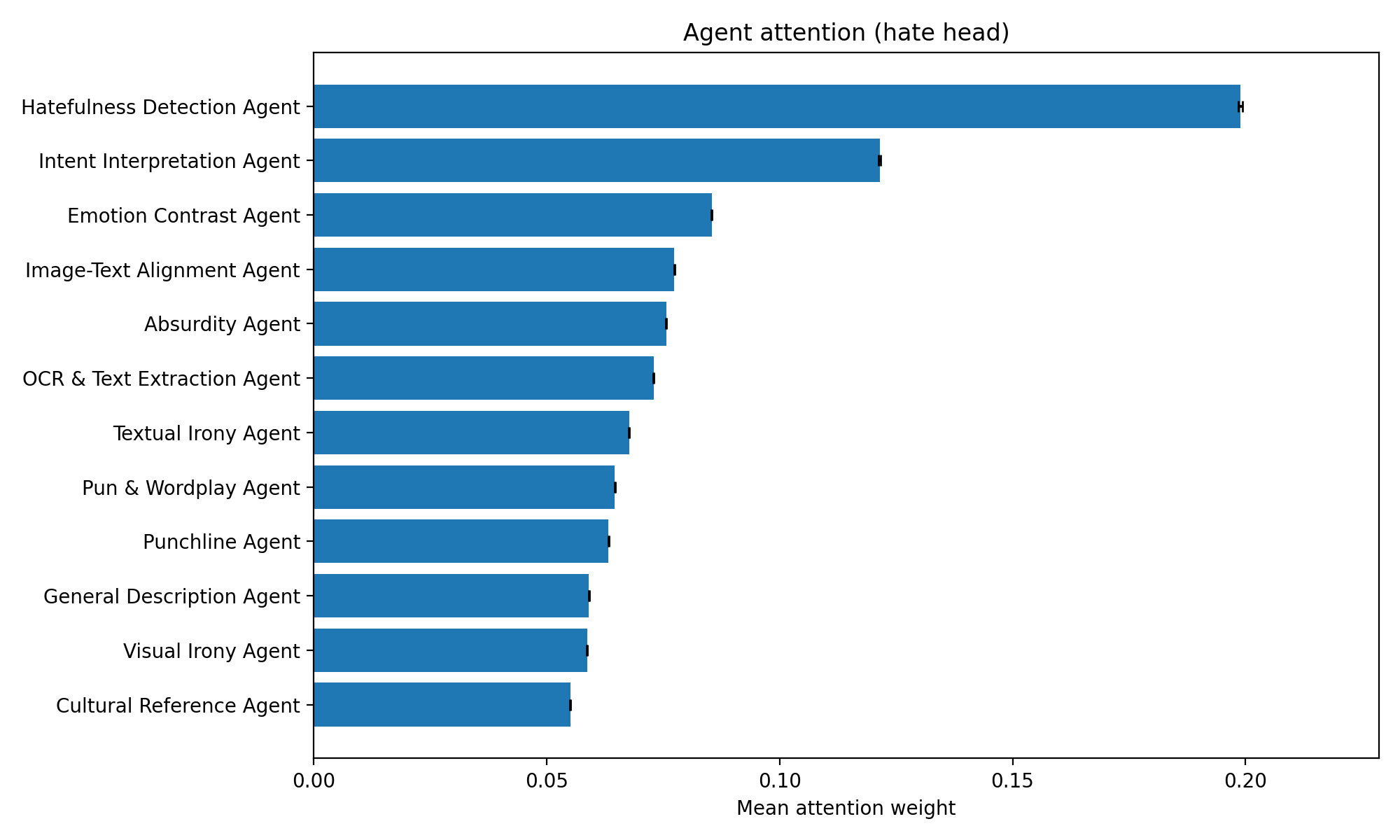}
  \caption{Ranked mean attention (with 95\% CI) over agents for the hate head. \emph{Hatefulness Detection} and \emph{Intent Interpretation} lead, followed by \emph{Emotion Contrast} and \emph{Image--Text Alignment}.}
  \label{fig:ranked-hate}
\end{figure}

\paragraph{Which agents matter most per task.}
Figures~\ref{fig:ranked-humour} and~\ref{fig:ranked-hate} present ranked mean attention with 95\% confidence intervals for the humour and hate heads, respectively. For the humour head, the model assigns the highest weight to \emph{Absurdity Agent} (mean $\approx 0.223$, CI $[0.215, 0.232]$), followed by \emph{OCR \& Text Extraction} (mean $\approx 0.125$, CI $[0.123, 0.127]$), with a long tail of humour-oriented agents such as \emph{Textual Irony}, \emph{Visual Irony}, \emph{Emotion Contrast}, and \emph{Cultural Reference}. For the hate head, \emph{Hatefulness Detection} (mean $\approx 0.199$, narrow CI around $[0.198, 0.199]$) and \emph{Intent Interpretation} (mean $\approx 0.121$) dominate, followed by \emph{Emotion Contrast} and \emph{Image--Text Alignment}. These patterns are consistent with the qualitative nature of the tasks: humour often hinges on textual setups and incongruity (captured by OCR and absurdity/irony agents), whereas hate classification prioritizes targeted derogation and intent.

\begin{figure}[t]
  \centering
  \includegraphics[width=\linewidth]{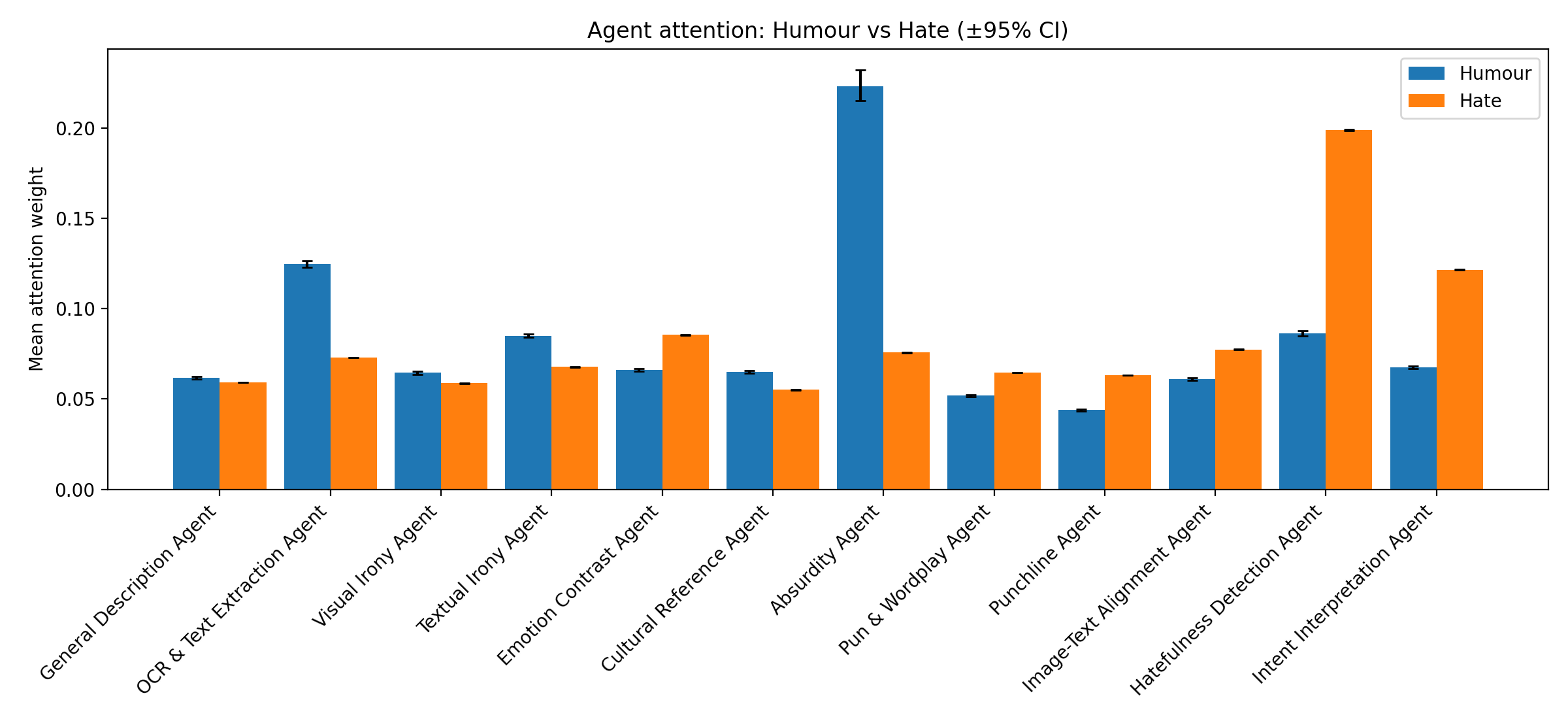}
  \caption{Side-by-side comparison of mean attention (with 95\% CI) for humour vs.\ hate across all agents. Humour places more weight on \emph{Absurdity} and \emph{OCR}; hate emphasizes \emph{Hatefulness Detection} and \emph{Intent Interpretation}.}
  \label{fig:compare}
\end{figure}

\begin{figure}[t]
  \centering
  \includegraphics[width=0.92\linewidth]{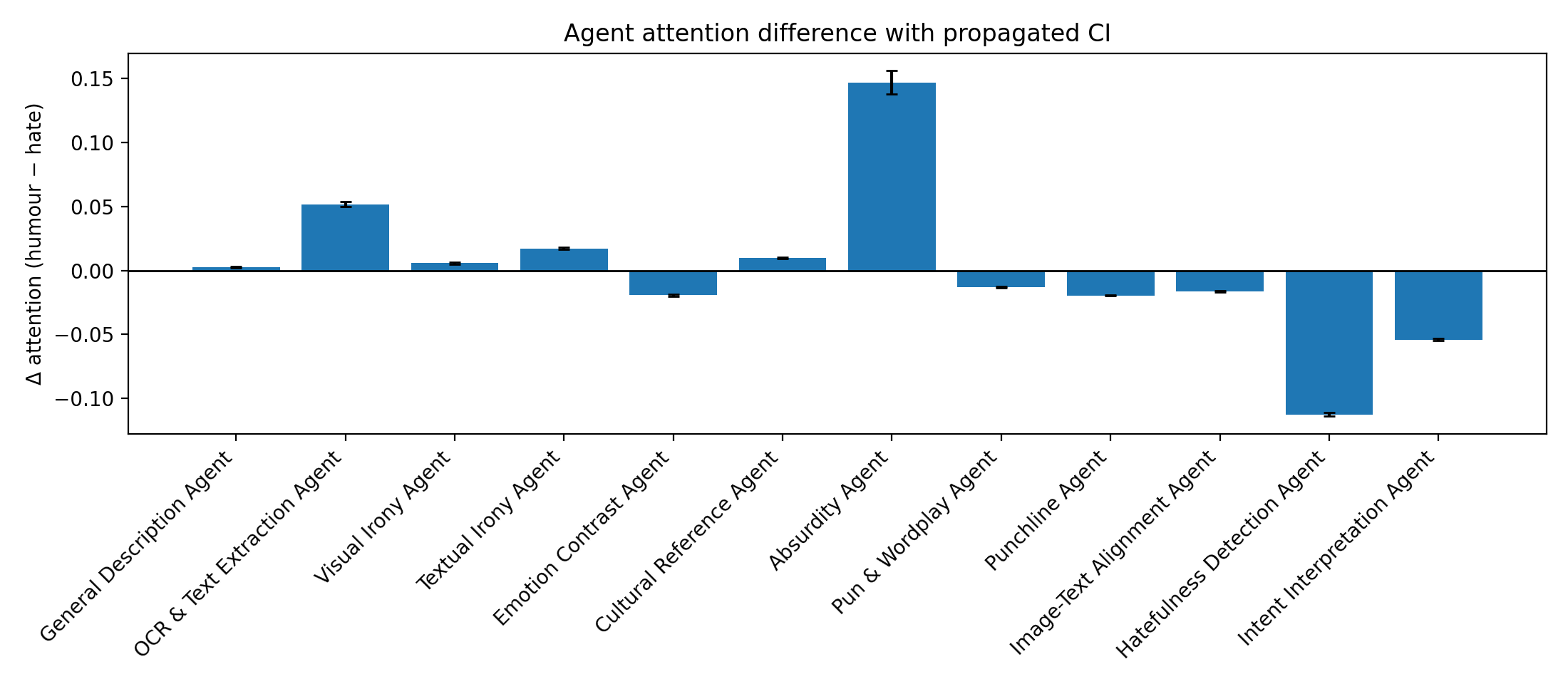}
  \caption{Attention differences (humour $-$ hate). Positive bars indicate higher weight under humour; negative bars indicate higher weight under hate. The largest positive deltas are for \emph{Absurdity} and \emph{OCR}; the largest negative deltas are for \emph{Hatefulness Detection} and \emph{Intent}.}
  \label{fig:delta}
\end{figure}

\paragraph{Humour and hate attend to different evidence.}
To directly compare tasks, Figure~\ref{fig:compare} shows side-by-side means (with 95\% CI) for each agent under the two heads, and Figure~\ref{fig:delta} plots the difference \emph{(humour $-$ hate)}. The largest positive deltas appear for \emph{Absurdity} and \emph{OCR}, confirming that humour is more text-anchored and playfully illogical. Conversely, the largest negative deltas occur for \emph{Hatefulness Detection} and \emph{Intent Interpretation}, indicating substantially higher emphasis under the hate head. These asymmetries provide quantitative support that the model relies on partially disjoint cues for the two tasks while still allowing overlap when a meme is simultaneously humorous and harmful.

\begin{figure}[t]
  \centering
  \includegraphics[width=0.9\linewidth]{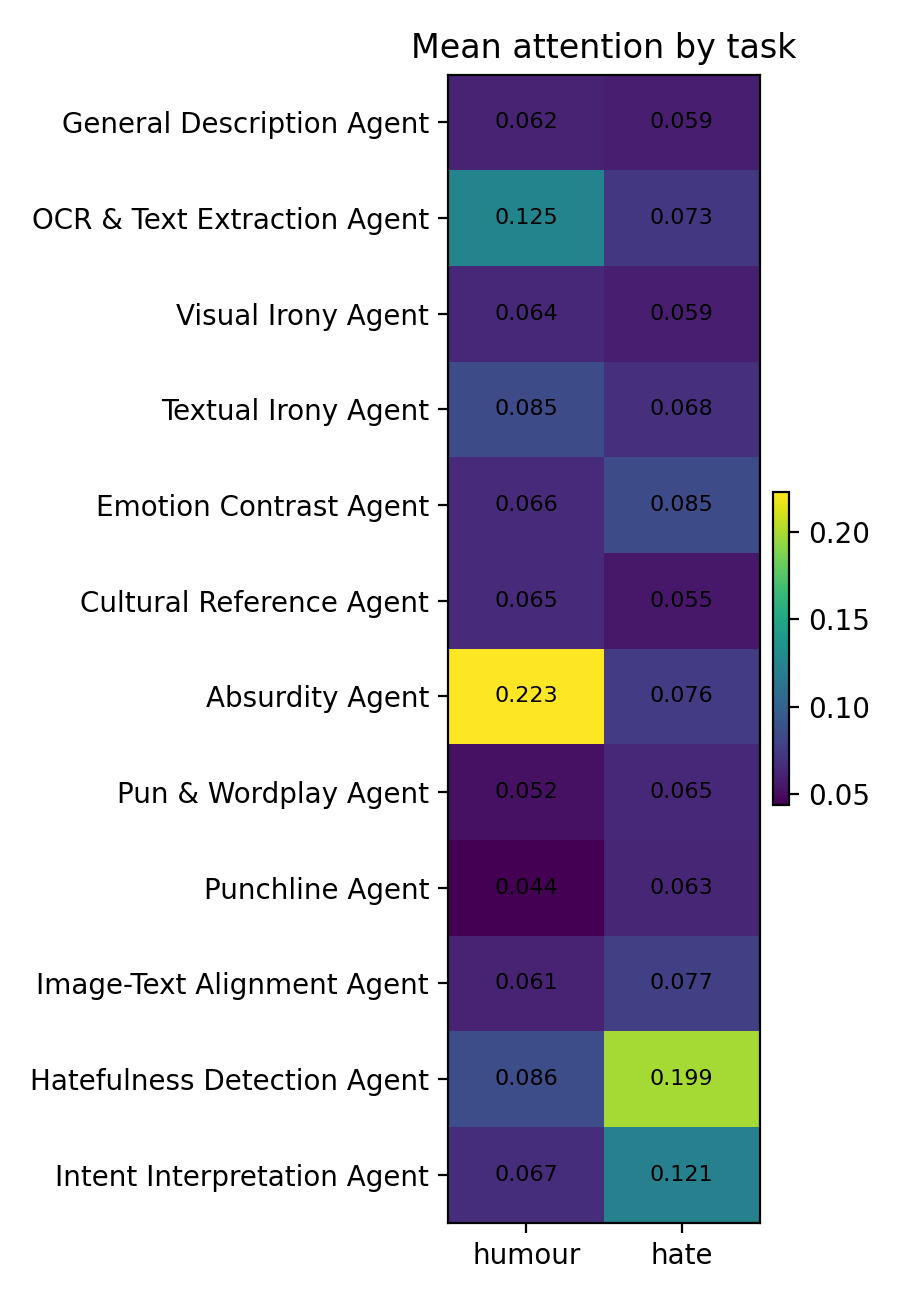}
  \caption{Heatmap of mean attention for each agent across \{humour, hate\}. The humour head is text- and absurdity-oriented; the hate head is safety- and intent-oriented.}
  \label{fig:heatmap}
\end{figure}

\paragraph{Compact overview across agents and tasks.}
Figure~\ref{fig:heatmap} provides a concise heatmap summarizing mean attention per agent across the two tasks. The humour head concentrates on \emph{Absurdity}, \emph{OCR}, and irony/contrast agents, whereas the hate head concentrates on \emph{Hatefulness Detection}, \emph{Intent}, and \emph{Alignment/Contrast}. This global view complements the ranked plots and makes the task-specific specialisation particularly transparent.

\begin{figure}[t]
  \centering
  \includegraphics[width=0.7\linewidth]{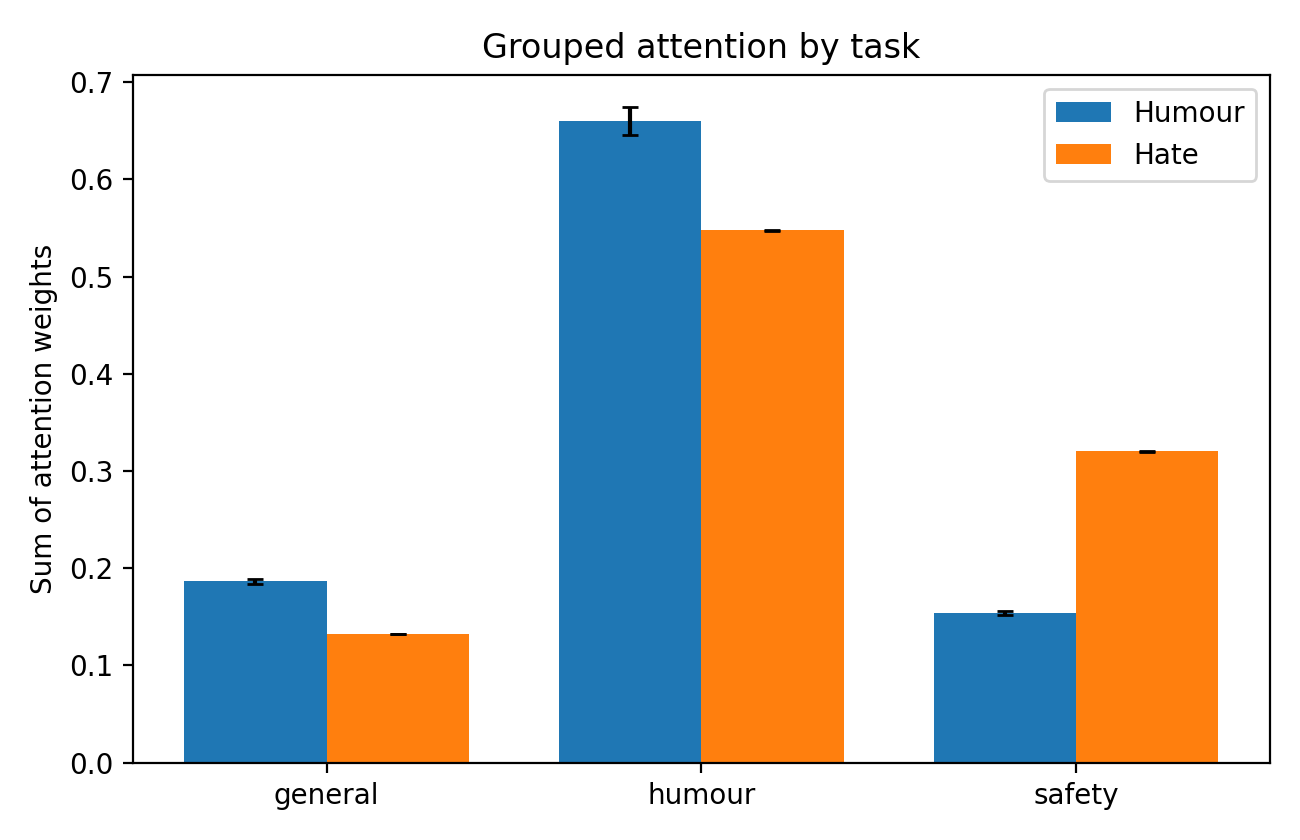}
  \caption{Grouped attention (mean with approximate CI) for \emph{general}, \emph{humour}, and \emph{safety} roles by task. The humour head emphasizes humour roles; the hate head emphasizes safety roles.}
  \label{fig:grouped}
\end{figure}

\paragraph{Group-level sanity and interpretability.}
Finally, Figure~\ref{fig:grouped} aggregates attention into three interpretable groups (\emph{general}, \emph{humour}, \emph{safety}). The humour head assigns the largest mass to the humour group, with non-trivial contributions from general cues (e.g., OCR and scene description). The hate head markedly increases the safety group’s share while still leveraging general and alignment/contrast information. This group-level analysis confirms that the learned attention respects intuitive priors and yields an interpretable division of labour across roles, supporting error analysis and governance use-cases.

\paragraph{Top-$5$ agents per task.}
For quick reference, Table~\ref{tab:top5} lists the top five agents for each task together with their mean attention and 95\% CI (rounded). These values are consistent with Figures~\ref{fig:ranked-humour}--\ref{fig:grouped} and further illustrate that the attention mechanism is learning task-aligned, human-interpretable signals.

\begin{table}[t]
\centering
\caption{Top-5 agents by mean attention for each task (mean $\pm$ 95\% CI).}
\label{tab:top5}
\resizebox{\linewidth}{!}{
\begin{tabular}{llcc}
\toprule
Task & Agent & Mean & 95\% CI \\
\midrule
Humour & Absurdity Agent & 0.223 & [0.215, 0.232] \\
       & OCR \& Text Extraction & 0.125 & [0.123, 0.127] \\
       & Hatefulness Detection & 0.086 & [0.085, 0.088] \\
       & Textual Irony & 0.085 & [0.084, 0.086] \\
       & Intent Interpretation & 0.067 & [0.067, 0.068] \\
\midrule
Hate   & Hatefulness Detection & 0.199 & [0.198, 0.199] \\
       & Intent Interpretation & 0.121 & [0.121, 0.122] \\
       & Emotion Contrast & 0.085 & [0.085, 0.085] \\
       & Image--Text Alignment & 0.077 & [0.077, 0.077] \\
       & Absurdity Agent & 0.076 & [0.076, 0.076] \\
\bottomrule
\end{tabular}}
\end{table}

\paragraph{Implications for novelty and research objectives.}
The attention distributions document that the model does not rely on a single undifferentiated feature space; instead, it adaptively emphasizes role-specialized evidence that is semantically aligned with each task. This is precisely the motivation for our multi-agent design and soft-attention aggregator. Moreover, the comparative analyses (Figures~\ref{fig:compare}--\ref{fig:delta}) empirically support the claim that humour and hate are not mutually exclusive but attend to different cues; the architecture can therefore capture co-occurrence by simultaneously up-weighting both humour- and safety-relevant agents when necessary. These findings reinforce the interpretability and governance value of our approach and substantiate the methodological contribution beyond accuracy-only baselines.

\section{Explanator–Agent Alignment and Faithfulness (Extended Analysis)}
\label{app:judge-faithfulness}
MAR-12 introduces explanation synthesizer that consumes (i) the 12 reasonings, (ii) the learned attention weights, and (iii) the system outcome, and then \emph{explains} the final decision \emph{in terms of} the attention-weighted agent evidence. The explanator’s job is not to discover attention but to faithfully narrate the model’s internal evidential path.

\paragraph{Metrics.}
Per image and task (humour, hate), we compute:
\begin{enumerate}
\item \textbf{Alignment} (\texttt{align\_humour}, \texttt{align\_hate}): textual alignment between the explanator's explanation and the 12 agents' reasonings, weighted by the model's learned attention.
\item \textbf{Margin} (\texttt{margin\_humour}, \texttt{margin\_hate}): confidence margin of the classifier's chosen label.
\item \textbf{Correctness} (\texttt{hum\_correct}, \texttt{hate\_correct}): indicator of prediction accuracy.
\end{enumerate}

\begin{table}[h]
\centering
\caption{Pearson correlations. Alignment captures faithfulness to agent reasoning/attention rather than correctness, whereas margin correlates with correctness as expected.}
\label{tab:faithfulness-corr}
\begin{tabular}{l r}
\toprule
\textbf{Pair} & \textbf{Pearson $r$} \\
\midrule
$\mathrm{align_{humour}}$ vs.\ $\mathrm{hum_{correct}}$ & $0.030$ \\
$\mathrm{align_{hate}}$ vs.\ $\mathrm{hate_{correct}}$ & $-0.026$ \\
$\mathrm{margin_{humour}}$ vs.\ $\mathrm{hum_{correct}}$ & $0.241$ \\
$\mathrm{margin_{hate}}$ vs.\ $\mathrm{hate_{correct}}$ & $0.267$ \\
\midrule
$\mathrm{align_{humour}}$ vs.\ $\mathrm{align_{hate}}$ & $0.425$ \\
$\mathrm{margin_{humour}}$ vs.\ $\mathrm{margin_{hate}}$ & $-0.053$ \\
$\mathrm{align_{humour}}$ vs.\ $\mathrm{margin_{humour}}$ & $-0.045$ \\
$\mathrm{align_{hate}}$ vs.\ $\mathrm{margin_{hate}}$ & $-0.033$ \\
\bottomrule
\end{tabular}
\end{table}

\paragraph{Key findings.}
\begin{itemize}
\item \textbf{Faithfulness is disentangled from correctness.}
Alignment is essentially uncorrelated with accuracy ($r\approx 0$ in both tasks), whereas the margin has the expected positive association with correctness (Table~\ref{tab:faithfulness-corr}). This is desirable: the explainer is evaluated on whether it \emph{follows} the attention-weighted agent evidence, not on whether the classifier is right. In contrast, the margin, as a confidence proxy, tracks correctness.

\item \textbf{Consistent explanatory style across tasks.}
The moderate correlation between humour and hate alignments ($r=0.425$) shows that MAR-12 induces a stable, role-aware explanatory pattern: when attention highlights certain agent roles, the explainer highlights the same roles irrespective of task.

\item \textbf{Margins reflect task difficulty, not shared structure.}
The near-zero correlation between humour and hate margins ($r=-0.053$) indicates decision difficulty varies idiosyncratically by task/image; this further argues against using margins to assess explanation quality.

\item \textbf{Cross-task reuse of humour signals on hate.}
On the hate/safety task, the attention module frequently allocates non-trivial mass to humour heads (harm in memes is often mediated via humour). The explainer mirrors this by citing humour-related cues as part of its hate/safety explanations, evidence that explanations reflect the \emph{same role mix} the model relied on.

\item \textbf{“Negative evidence” behavior.}
When humour is absent, the attention module still consults humour heads to \emph{verify lack of signal}; the explainer correspondingly points to those heads to justify a “not humorous” outcome. Explanations thus remain faithful even for negative conclusions, not only for positive attributions.

\item \textbf{Stability across label quadrants.}
Attention distributions (especially for hate) remain similar across HUM/HATE label combinations, and the explainer’s emphasis follows this stability, ruling out label-chasing and supporting that the explainer narrates the attention-determined process rather than the label itself.
\end{itemize}

\begin{table}[h]
\centering
\caption{Descriptive statistics for alignment and margin (per image).}
\label{tab:faithfulness-desc}
\begin{tabular}{l r r r r}
\toprule
\textbf{Metric} & \textbf{Mean} & \textbf{Std} & \textbf{Min} & \textbf{Max} \\
\midrule
$\mathrm{align\_humour}$ & $0.1123$ & $0.0375$ & $0.00019$ & $0.36440$ \\
$\mathrm{align\_hate}$ & $0.1284$ & $0.0395$ & $0.00020$ & $0.24527$ \\
$\mathrm{margin\_humour}$ & $0.7018$ & $0.2642$ & $0.00130$ & $0.98738$ \\
$\mathrm{margin\_hate}$ & $0.6139$ & $0.2717$ & $0.00019$ & $0.97455$ \\
\bottomrule
\end{tabular}
\end{table}

\paragraph{Interpretation of magnitudes.}
Alignment values are concentrated in a narrow band (means $\approx$0.11–0.13; std $\approx$0.04; Table~\ref{tab:faithfulness-desc}), indicating \emph{consistent grounding} of the explainer’s text in attention-weighted agent rationales across items. By contrast, margins vary widely (std $\approx$0.26–0.27), reflecting heterogeneous decision difficulty—another reason alignment should not be conflated with correctness.

\paragraph{Why this is novel in MAR-12.}
\begin{enumerate}
\item \textbf{Decoupled selection–narration.}
By learning attention \emph{outside} the explainer and requiring the explainer to condition on those weights, MAR-12 makes faithfulness \emph{testable}. Our results show the explainer’s narrative tracks the external, role-aware attention rather than inventing its own saliency.
\item \textbf{Role-aware, multi-agent alignment.}
Explanations reflect domain structure (humour, safety, general, \dots) and naturally capture cross-task coupling (humour cues shaping hate decisions) when attention says they matter.
\item \textbf{Faithfulness for negative conclusions.}
The system surfaces \emph{absence of evidence} via the same roles that would have been diagnostic, a property often missing in post-hoc LLM explanations.
\item \textbf{Separation of concerns in evaluation.}
Alignment measures \emph{faithfulness} (process-level grounding), while margin measures \emph{confidence/accuracy}. The empirical orthogonality between them is a distinctive empirical signature of our design.
\end{enumerate}

\paragraph{Implications.}
MAR-12 yields explanations that are \emph{causally grounded} in the model’s internal evidence flow (agents $\rightarrow$ attention $\rightarrow$ classifier), independent of outcome accuracy, while preserving the intuitive link between confidence and correctness. This supports transparent, accountable meme understanding: reviewers can inspect \emph{which} roles the model relied on (attention) and \emph{how} the explainer’s text follows that evidence (alignment).

\end{document}